\documentclass[10pt,twocolumn,letterpaper]{article}

\usepackage{iccv}
\usepackage{times}
\usepackage{epsfig}
\usepackage{graphicx}
\usepackage{amsmath}
\usepackage{amssymb}
\usepackage{multirow}
\usepackage{mathrsfs}
\usepackage[T1]{fontenc}
\usepackage{graphicx}
\usepackage{array}
\usepackage{mdwmath}
\usepackage{mdwtab}
\usepackage{eqparbox}
\usepackage{fixltx2e}
\usepackage{url}
\usepackage{textcomp}
\usepackage{lipsum}
\usepackage{multirow}
\usepackage{subcaption}
\usepackage{amsfonts,amssymb,amsthm}
\usepackage{xspace}
\usepackage{xcolor}
\usepackage{bm}
\usepackage{bbm}
\usepackage{booktabs}
\usepackage{etoolbox}
\usepackage{makecell}
\usepackage{nicefrac}
\usepackage{textcomp}
\usepackage{stmaryrd}
\usepackage{mathrsfs}
\usepackage{paralist}
\usepackage{comment}

\usepackage[font=footnotesize,labelsep=space,labelfont=bf]{caption}

\usepackage{url}

\makeatletter

\newcommand{\Rmnum}[1]{\expandafter\@slowromancap\romannumeral #1@}
\makeatother

\newtheorem*{theorem*}{Theorem}
\newtheorem*{lemma*}{Lemma}
\newtheorem*{cor*}{Corollary}

\newcommand{\namedref}[2]{\hyperref[#2]{#1~\ref*{#2}}}

\newcommand{\Algorithmref}[1]{\namedref{Algorithm}{algo:#1}}

\newcommand{\Sectionref}[1]{\namedref{Section}{sec:#1}}
\newcommand{\Subsectionref}[1]{\namedref{Section}{subsec:#1}}

\newcommand{\Appendixref}[1]{\namedref{Appendix}{app:#1}}

\newcommand{\Tableref}[1]{\namedref{Table}{tab:#1}}
\newcommand{\Figureref}[1]{\namedref{Figure}{fig:#1}}

\newcommand{\Pageref}[1]{\hyperref[#1]{page~\pageref*{#1}}}

\definecolor{darkred}{rgb}{0.5, 0, 0} 
\definecolor{darkblue}{rgb}{0,0,0.5}

\newcommand{\calL}{\ensuremath{\mathcal{L}}\xspace}

\newcommand{\R}{\ensuremath{\mathbb{R}}\xspace}

\renewcommand{\paragraph}[1]{\smallskip\noindent{\bf #1}~}

\usepackage{mathtools}

\usepackage{algorithm}
\usepackage{algorithmicx}
\usepackage{algpseudocode}

\usepackage[accsupp]{axessibility}  
\algdef{SE}[DOWHILE]{Do}{doWhile}{\algorithmicdo}[1]{\algorithmicwhile\ #1}%

\usepackage{enumitem}
\usepackage{float}

\usepackage[pagebackref=true,breaklinks=true,letterpaper=true,colorlinks,bookmarks=false]{hyperref}

\iccvfinalcopy 

\def\iccvPaperID{11613} 

\ificcvfinal\fi

\begin{document}

\title{\textsc{Sempart}: Self-supervised Multi-resolution Partitioning of Image Semantics}
\author{Sriram Ravindran\\
Adobe\\
{\tt\small sravindr@adobe.com}
\and
Debraj Basu\\
Adobe\\
{\tt\small dbasu@adobe.com}
}

\twocolumn[{
\renewcommand\twocolumn[1][]{#1}
\maketitle
\begin{center}
\vspace{-1mm}
  \includegraphics[width=1\linewidth]{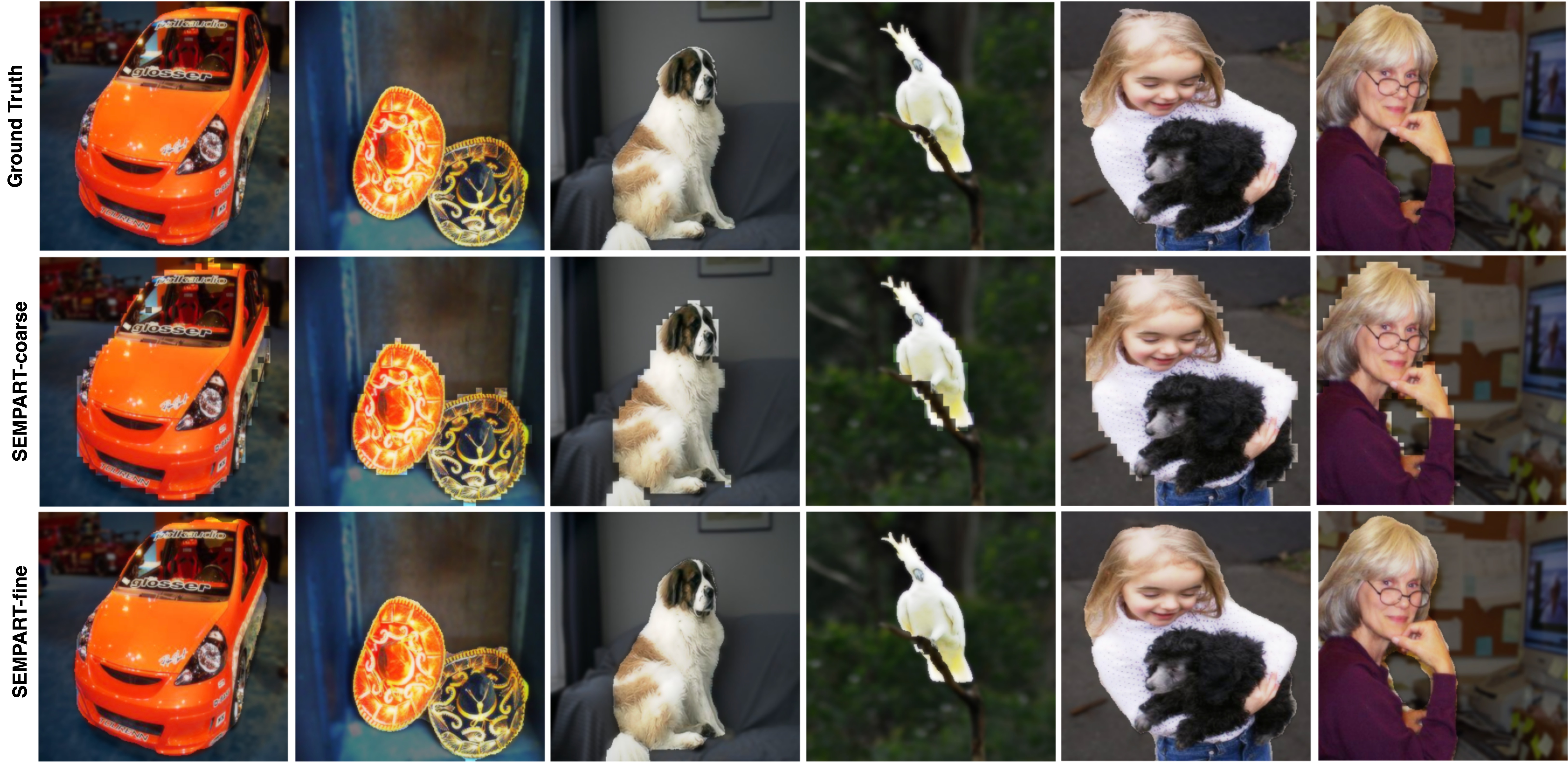}
  \captionsetup{type=figure,font=small}
  \vspace{-0.5cm}
  \caption{Each image is the original image with an overlayed saliency mask. The first row of every column utilizes the ground truth saliency mask, the second and third rows overlay with the self-supervised \textsc{Sempart}-coarse and -fine masks on the same image respectively.}
  \label{fig:teaser}
\vspace{-0.1cm}
\end{center}
}]

\maketitle
\ificcvfinal\thispagestyle{empty}\fi

\begin{abstract}
\vspace{-0.3cm}Accurately determining salient regions of an image is challenging when labeled data is scarce. DINO-based self-supervised approaches have recently leveraged meaningful image semantics captured by patch-wise features for locating foreground objects. Recent methods have also incorporated intuitive priors and demonstrated value in unsupervised methods for object partitioning. In this paper, we propose \textsc{Sempart}, which jointly infers coarse and fine bi-partitions over an image's DINO-based semantic graph. Furthermore, \textsc{Sempart} preserves fine boundary details using graph-driven regularization and successfully distills the coarse mask semantics into the fine mask. Our salient object detection and single object localization findings suggest that \textsc{Sempart} produces high-quality masks rapidly without additional post-processing and benefits from co-optimizing the coarse and fine branches.
\end{abstract}

\section{Introduction}
\label{sec:intro}
\begin{figure}
\begin{center}
\scalebox{0.95}{
\includegraphics[width=0.5\textwidth]{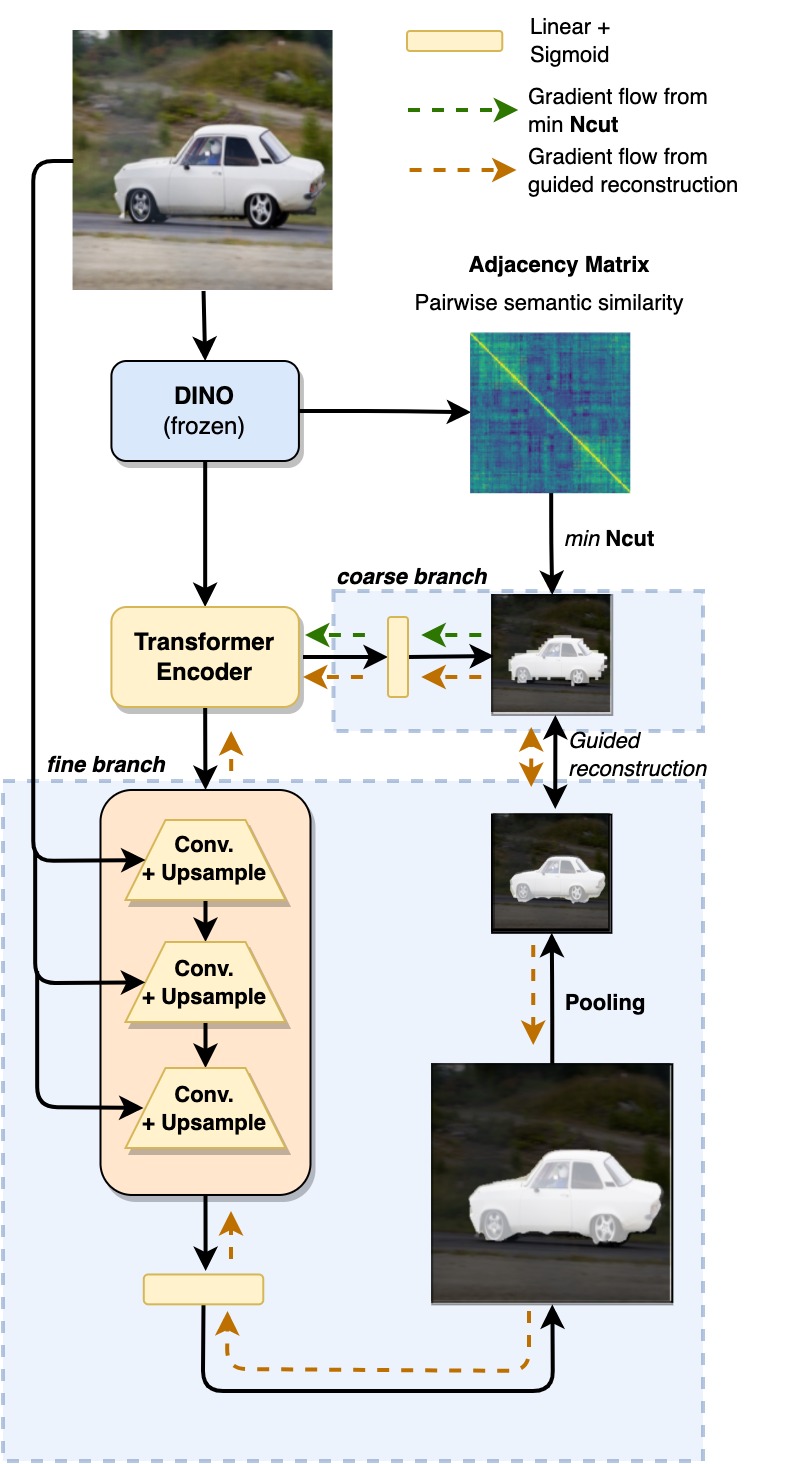}}
\end{center}\vspace{-0.5cm}
\caption{Overview of \textsc{Sempart}: We refine the SSL features into co-optimized low resolution {\it coarse} and high resolution {\it fine masks}, based on graph cut and guided super-resolution respectively.}\vspace{-0.5cm}
\label{fig:sempart}
\end{figure}
Identifying salient regions of an image prone to holding visual attention remains a long-standing fuzzy problem~\cite{fuzzy} relying significantly on carefully annotated data~\cite{sod_survey1,sod_survey2,tokencut}. Recently self-supervised (SSL) mechanisms based on large-scale pre-trained backbones~\cite{moco,swav,mae}, such as DINO~\cite{dino}, have demonstrated increased capability in segmenting images~\cite{stego,deepspectralclustering} and extracting objects in the foreground~\cite{lost,selfmask,tokencut,move,found}. 

The unavailability of labels is limiting to inferring high-quality object masks. However, many recent methods have demonstrated that incorporating well-informed priors into the partitioning process is significantly beneficial to finding saliency regions and foreground objects in an unsupervised setting~\cite{iem,lost,gen1, gen2, gen3,tokencut,move,selfmask}.

Different forms of statistical independence of the foreground have driven recent approaches, with the most recent state-of-the-art focusing on movability~\cite{move} of the salient object. Distinguishability and predictability of the foreground from the background have also been successful indicators.
For example, statistical variations such as in color and texture of the foreground minimally alter the overall distribution of the population~\cite{redraw}. Furthermore, in-painting models such as MAE~\cite{mae} have been particularly effective at measuring predictability~\cite{iem} and defining movability~\cite{move}. 

Inferring graph signals~\cite{gsp} for partitioning a semantic graph over an image has gained popularity~\cite{selfmask,tokencut,lost,deepspectralclustering,deepcut}, with recent methods establishing surprisingly strong baselines using traditional techniques. In particular, the solution to the relaxation of the NP-complete discrete normalized cut problem~\cite{normcut} first demonstrated promise in unsupervised image segmentation, which has further translated to recent findings in \cite{selfmask,tokencut,namedmask}. 

\cite{dlncut1,dlncut2} discuss the benefit of learning to predict spectral decomposition for a graph and employ graph neural networks in a reinforcement learning setup for predictively performing the normalized cut. More recently, \cite{weaksupervisedcnn} leveraged normalized cut for regularizing a convolutional network driven by partial cross entropy loss in a weakly supervised setting and demonstrated significant performance improvement. More broadly, spectral partitioning of semantic graphs~\cite{selfmask,tokencut,deepspectralclustering,deepcut} has become an emerging underlying theme for detecting salient regions.

\noindent{\bf Contributions. }In this paper, we propose \textsc{Sempart}, which builds on ideas from \cite{tokencut,pixtransform,guidedsuperresolution} for producing high-quality foreground masks in an SSL setting. \textsc{Sempart} learns a transformer-based encoder that refines the patchwise DINO features for inferring a relaxation of graph cut that minimizes the expected normalized cut loss~\cite{dlncut1} over a semantic graph informed by DINO feature correspondences. 

As seen in \cite{tokencut,lost,dino}, the foreground masks obtained abandon the fine boundary details from processing features at a low resolution.
Unlike \cite{selfmask,tokencut,lost}, which perform successive refinement of the {\it coarse masks} post-inference, \textsc{Sempart} implements a convolutional {\it fine branch} that processes and supplements the transformed DINO features with RGB features at progressively increasing resolutions for producing original resolution {\it fine masks}. Motivated by \cite{pixtransform,guidedsuperresolution}, \textsc{Sempart} treats the {\it coarse mask} as the source and the image as a guide for inferring high-quality {\it fine masks} (see \Figureref{teaser}, \Tableref{main_results}) regularized by weighted neighborhood-based graph total variation~\cite{gtv}. 

\noindent In summary, our contributions are as follows:
\begin{itemize}
\item We propose a novel strategy for co-optimizing {\it coarse} and {\it fine masks}, that decouples image partitioning into semantic separation of rich self-supervised features and high-frequency detailing, respectively.
    
    \item \textsc{Sempart} outperforms recent state-of-the-art methods in saliency detection by 3.7\% in $\max F_\beta$ and 2.7\% in IoU on average and emits high-quality bounding boxes for locating objects.
    \item \textsc{Sempart} produces high-quality {\it fine masks} rapidly by eliminating time-consuming post-inference iterative refinement and saving 200ms on average.
    
\end{itemize}

\section{Related work}
\label{sec:related_work}
Vision systems have historically benefited from segmenting a scene into objects constituting salient regions~\cite{sod_survey1}. Supervised mechanisms~\cite{u2net,selfreformer} have dominated the landscape despite the prohibitive costs of obtaining labeled data. Traditional unsupervised approaches~\cite{sod_survey2,relworktrad1} have encoded beliefs about the foreground region, such as differences in color and contrast and objectness and depth perception, into partitioning techniques.

{\bf Spectral methods.} Graph-based techniques have received interest wherein spectral partitioning is undertaken over a graphical representation of an image deduced from the priors. \cite{normcut} proposed {\it normalized cut} as an improvement over the {\it min cut} criterion~\cite{mincut}, for producing clusters that are well balanced. The relaxation of the discrete problem involved a spectral analysis of the symmetrically normalized graph Laplacian \begin{align}
    L=I-D^{-\nicefrac{1}{2}}WD^{-\nicefrac{1}{2}}.
\end{align}The unavailability of effective semantic similarity measures between regions of an image for populating the adjacency matrix $W$ inhibited the quality of resulting partitions. 

{\bf Self-supervised representations.} With the emergence of deep techniques for learning contextually aware representations~\cite{dino, mae,moco,swav},many of these traditional prior-based techniques have demonstrated increased effectiveness and therefore received renewed interest. The semantically aware DINO~\cite{dino} features were used for implementing seed expansion into salient regions, initialized with patches that are least similar to other patches as seed in LOST~\cite{lost}. On the contrary, FOUND~\cite{found} locates a background seed first and then expands it. In \cite{freesolo}, a SOLO~\cite{solo} model is trained on coarse masks extracted using SSL features, for instance segmentation. 

The memory bottleneck of attention mechanism~\cite{bert} prevents low-resolution deep SSL features from capturing high-frequency details of an image which often are only helpful in predicting coarse masks~\cite{lost,selfmask, tokencut}. Therefore, despite significant performance gains, these methods require computationally heavy post-processing~\cite{bilateralfiltering,crf,crf2} to generate high-quality fine masks.

{\bf Inpainting} as a helpful object detection tool was first proposed in \cite{iem}, which hypothesized that it is difficult to predict the foreground given a background and vice versa. SSL features from masked autoencoder (MAE~\cite{mae}) were also leveraged by recent state-of-the-art MOVE~\cite{move} for adversarially training a convolutional mask generator for distinguishing between real- and fake-inpainted images based on movability of salient objects. MOVE established superiority in detecting both salient regions as well as single objects. The movability criterion allows MOVE to directly predict saliency masks at a high resolution which is also why it outperformed its counterparts without post-processing.

\textsc{SelfMask}~\cite{selfmask} uses multi-model SSL features~\cite{dino, moco,swav} for populating $W$ and constructs pseudo ground truth saliency masks for a subsequent MaskFormer~\cite{maskformer} training by clustering eigenvectors of the unnormalized graph Laplacian. Along similar lines, \cite{deepspectralclustering} employs clustering based on normalized Laplacian for semantic segmentation and object localization. 

Our work is most closely related to \cite{tokencut,dlncut1,pixtransform,guidedsuperresolution}. TokenCut~\cite{tokencut} makes the bi-partitioning mathematically precise by using the eigenvector with the second smallest eigenvalue, which corresponds to a relaxation of the normalized cut~\cite{normcut} problem and demonstrates value in pursuing graph-based techniques for detecting salient regions. 

Iterative computations during inference with expensive post-processing~\cite{tokencut,deepspectralclustering}, or otherwise training in two stages leveraging multiple SSL models~\cite{selfmask} for improving performance, can be limiting. To alleviate this, we follow MOVE's approach of training a single bi-partitioning model as a transformation of the DINO backbone (see \Figureref{sempart}) and encode our novel strategies into the loss functions (see \Subsectionref{sempart}, \Subsectionref{gtv}). The \textsc{Sempart} architecture involves a {\it fine branch} inspired by graph-driven iterative techniques for super-resolution~\cite{pixtransform,guidedsuperresolution} for predicting accurate high-resolution masks.

Minimizing expected graph cut losses over a population was previously evaluated in \cite{dlncut1,dlncut2,deepcut}, which proposed to optimize expected normalized cut using graph neural networks. We show that \textsc{Sempart} exhibits similar benefits (see \Tableref{main_results}, \Tableref{singleobject}) from jointly inferred graph-driven bi-partitioning and graph regularized guided super-resolution for generating high-fidelity saliency masks rapidly without any post-processing or multi-stage training.

\section{Approach}
\label{sec:approach}
In this work, we detect salient regions and localize single objects within an image by learning to partition the image into two regions that are semantically less related~\cite{tokencut,stego,selfmask,deepcut}. We leverage DINO~\cite{dino}, which provides effective pre-trained SSL feature correspondences~\cite{tokencut,stego,deepspectralclustering} for learning a {\it coarse} binary mask that partitions a semantic graph constructed between image patches as nodes. Motivated by image-guided super-resolution~\cite{pixtransform} and graph regularization~\cite{guidedsuperresolution,gtv}, we co-optimize and infer masks at the original resolution in parallel, thereby correcting a {\it coarse mask's} inaccuracies, preserving fine boundary details.

\subsection{Background}
\label{subsec:background}
\noindent {\bf Normalized Cut.}
The normalized cut~\cite{normcut} of a weighted undirected complete graph $G=(V,E,w)$ where $w_{ij}>0$ denotes the weight of $(i,j)\in E$, is given by a binary graph signal $s:v\in V\rightarrow s(v)\in\{0,1\}$ that minimizes 
\begin{align}
\text{Ncut}(A,B)=\frac{w(A,B)}{w(A,V)} + \frac{w(B,A)}{w(B,V)}\label{ncut1}
\end{align}
where $A\coloneqq \{v|v\in V,s(v)=0\}$, $B\coloneqq \{v|v\in V,s(v)=1\}$ and $w(A,B)\coloneqq\sum_{s(i)=0,s(j)=1}w_{i,j}$. 

\vspace{2mm}

Being NP-complete, Shi et al.~\cite{normcut} first proposed to solve a relaxation which amounts to solving a generalized eigensystem followed by discretization. More recently, the relaxation of \eqref{ncut2} has been effective at semantically segmenting images in a self-supervised manner~\cite{tokencut}. Motivated by \cite{dlncut1,dlncut2}, non-linear parameterizations of the graph signal have enabled deep partitioning~\cite{deepcut} and regularization~\cite{weaksupervisedcnn} based on normalized cut.

\noindent {\bf Deep self-supervised feature correspondences.}
Large-scale pre-trained self-supervised image embedders such as DINO~\cite{dino}, MAE~\cite{mae}, MoCo~\cite{moco}, SwAV~\cite{swav} possess beneficial emergent properties for downstream tasks~\cite{lost, tokencut, move, selfmask,stego}. These models are based on vision transformers~\cite{vit}, which generate an embedding for each patch. Specifically, given an image of dimensions $C\times H\times W$, and an SSL embedder operating with patch size $p$, we obtain a tensor of size $\mathit{D\times (H/p\times W/p + 1)}$, including the embedding for the \texttt{[CLS]} token that represents the entire image. In this paper, we leverage DINO as it emits semantically relevant embeddings~\cite{dino, tokencut, lost, found,stego}.

In particular, \cite{tokencut} computed an affinity matrix using the feature correspondences from DINO. A graph view of the output is considered where the graph $G = (V, E)$ contains patches $V$, and connections between any two patches are encoded in the edge list $E$. Each patch $v \in V$ has an associated normalized DINO embedding $F_{v}$. The affinity matrix is given by the feature correspondences, 
\begin{align}
W_{ij} = \left\{
\begin{array}{l}
    1 \mid \langle F_{v_i},F_{v_j}\rangle > \tau \\
    \epsilon \mid otherwise.
\end{array}\label{ncutaff}
\right.
\end{align}

\subsection{Self-supervised multi-resolution partitioning ({\bf \textsc{Sempart}})}
\label{subsec:sempart}
We propose \textsc{Sempart}, which converts an image into a semantic graph $G$ over non-overlapping patches, which form the set of nodes $V$. \textsc{Sempart}'s architecture (see \Figureref{sempart}) has two main branches that infer a {\it coarse} and {\it fine mask} jointly, which are informed by normalized cut and image-guided super-resolution, respectively.
We posit that guided super-resolution not only refines the {\it coarse mask} into a {\it fine mask} by preserving high-resolution details. It also helps regularize the overall learning and justifies our co-optimization strategy.

\vspace{1mm}
\noindent{\bf Normalized cut for \textit{coarse mask}.}
\renewcommand{\arraystretch}{1.2}
\begin{table*}
\begin{center}
\scalebox{0.95}{
\begin{tabular}{c|l|c|c|c|c|c|c|c|c|c}
\hline
&\multirow{2}{*}{\textbf{\parbox{4.5cm}{Model}}} & 
\multicolumn{3}{c|}{\textbf{DUT-OMRON~\cite{dutomron}}} & 
\multicolumn{3}{c|}{\textbf{DUTS-TE~\cite{duts}}} & 
\multicolumn{3}{c}{\textbf{ECSSD~\cite{ecssd}}} \\
\cline{3-11}
&& \textbf{Acc} & \textbf{IoU} & $\mathbf{max F_\beta}$ & \textbf{Acc} & \textbf{IoU} & $\mathbf{max F_\beta}$ & \textbf{Acc} & \textbf{IoU} & $\mathbf{max F_\beta}$ \\
\hline
\parbox[t]{2mm}{\multirow{6}{*}{\rotatebox[origin=c]{90}{Method}}}&LOST~\cite{lost}  & .797  & .410 & .473 & .871 & .518 & .611 & .895 & .654 & .758 \\
&TokenCut~\cite{tokencut}  & .880 &  .533 & .600 & .903 & .576 & .672 & .918 & .712 & .803 \\
&FreeSOLO~\cite{freesolo}  & .909 & .560 &  .684 & .924 & .613 & .750 &  .917 & .703 & .858 \\
&MOVE~\cite{move} & .923 & .615 & .712 & .950 & .713 & .815 & .954 & .830 & .916 \\
&\textbf{\textsc{Sempart}-Coarse}  & \textbf{.932}  & .640 & .755 & .956 & .727 & .864 & .961 & .837 & .943 \\
&\textbf{\textsc{Sempart}-Fine} & \textbf{.932} & \textbf{.668} &\textbf{.764} & \underline{\textbf{.959}} & \underline{\textbf{.749}} & \textbf{.867} & \textbf{.964} & \textbf{.855} & \underline{\textbf{.947}} \\
\hline
\parbox[t]{2mm}{\multirow{5}{*}{\rotatebox[origin=c]{90}{+ BF}}}&LOST+BF & .818 & .489 & .578 & .887 & .572 & .697 & .916 & .723 & .837 \\
&TokenCut+BF & .897 & .618 & .697 & .914 & .624 & .755 & .934 & .772 & .874 \\
&MOVE+BF & .931 & .636 & .734 & .951 & .687 & .821 & .953 & .801 & .916 \\
&\textbf{\textsc{Sempart}-Coarse+BF} & \textbf{.934} & \textbf{.661} & \textbf{.764} & \textbf{.957} & \textbf{.697} & \textbf{.858} & \textbf{.960} & \textbf{.820} & \textbf{.932} \\
&\textbf{\textsc{Sempart}-Fine+BF} & .933 & .653 & .760 & .955 & .685 & .853 & .959 & .816 & .931 \\
\hline
\parbox[t]{2mm}{\multirow{5}{*}{\rotatebox[origin=c]{90}{+ \textsc{SelfMask}}}}&\textsc{SelfMask} on pseudo + BF~\cite{selfmask}  & .919 & .655 & (.774)\textsuperscript{*} & .933 & .660 & (.819)\textsuperscript{*} & .955 & .818 & (.911)\textsuperscript{*} \\
&\textsc{SelfMask} on MOVE & .933 & .666 & .756 & .954 & .728 & .829 & .956 & .835 & .921 \\
&\textsc{SelfMask} on MOVE + BF & .937 & .665 & .766 & .952 & .687 & .827 & .952 & .800 & .917\\
&\textbf{\textsc{SelfMask} on \textsc{Sempart}-Coarse} & .936 & .675 & .773 & \textbf{.958} & .743 & .872 & .962 & .843 & .938 \\
&\textbf{\textsc{SelfMask} on \textsc{Sempart}-Fine} & \underline{\textbf{.942}} & \underline{\textbf{.698}} & \underline{\textbf{.799}} & \textbf{.958} & \underline{\textbf{.749}} & \underline{\textbf{.879}} & \textbf{.963} & \textbf{.850} & \textbf{.944} \\
\hline

\parbox[t]{2mm} & U\textsuperscript{2}-Net (supervised)  & .928 & .693 & .771 & .943 & .733 & .822 & \underline{\textbf{.967}} & \underline{\textbf{.878}} & \underline{\textbf{.947}} \\
\hline

\end{tabular}}
\end{center}\vspace{-0.3cm}
\scriptsize\textsuperscript{*} The $\max F_\beta$ for \textsc{SelfMask} on pseudo + BF is reported within brackets (), from the reevaluation in \cite{move} upon confirming that it was originally calculated incorrectly~\cite{move,found}.\vspace{-0.3cm}
\caption{Quantitative comparison of \textsc{Sempart} with state-of-the-art MOVE and other related works for saliency detection. \textsc{Sempart}-Coarse and -Fine outperform MOVE significantly in all three evaluation categories (Method, +BF, +\textsc{SelfMask}) across all datasets. The best-performing method in a category and across categories is in {\bf bold} and {\bf \underline{underlined}}, respectively.}
\vspace{-0.5cm}
\label{tab:main_results}
\end{table*}
A frozen DINO backbone transforms the input image $X\in\R^{3\times320\times320}$ into low-resolution SSL features $F\in\R^{64\times40\times 40}$.
We apply a single layer transformer encoder with two attention heads, followed by a {\it coarse branch} (see \Figureref{sempart}) comprised of a linear classification head, for transforming the low resolution features into a {\it coarse} saliency mask in the form of a soft partitioning indicator vector $S_{\text{coarse}}\in[0,1]^{|V|}$ where $|V|=40\times 40$. For partitions A and B with their indicator vectors $S_A=S_{\text{coarse}}$ and $S_B=1-S_A$, \eqref{ncut1} is rewritten as
\begin{align}
    \calL_{\text{Ncut}}(X)\coloneqq\text{Ncut}(A,B)=\sum_{i\in\{A,B\}}\frac{S_i^TW(1-S_i)}{{S_i}^TW\mathbf{1}}.\label{ncut2}
\end{align}
This results in a {\it coarse mask} at $40\times40$, which amplifies the semantic distinguishability between the two partitions where the affinity between image patches $i$ and $j$ is computed using the DINO embeddings in \eqref{ncutaff} and denoted by $W_{ij}$. Upon minimizing this heuristic over the entire population, we see a significant improvement in performance over solving the generalized eigensystem in \cite{tokencut} (see \Tableref{main_results}). 

\vspace{1mm}
\noindent{\bf Guided super-resolution for \textit{fine mask}.}
The generated {\it coarse mask} often fails to capture finer high-frequency details~\cite{tokencut,pixtransform} at the original image resolution, which is detrimental to the performance in detecting salient regions. Previously, such methods have employed expensive iterative post-processing such as Bilateral Filtering~\cite{bilateralfiltering, selfmask, lost, tokencut} or CRF~\cite{crf, stego} for every inferenced image. These methods utilize pixels' color and positional information to readjust the generated {\it coarse masks}. The possibility of erosion of the mask has been discussed as a limitation in \cite{move}.

By delegating the generation of linearly separable semantic features to the {\it coarse branch}, our architecture enables a refinement network to exclusively focus on detailing and denoising at higher frequencies and around the edges.
We jointly optimize a {\it fine branch} (see \Figureref{sempart}) comprised of a convolutional mask refinement network inspired by a recent guided super-resolution technique~\cite{pixtransform} which trains a multi-layer perceptron for enhancing the mask with guidance from the image. While \cite{pixtransform} performs iterative refinement per image, we co-optimize our refinement network for predicting a {\it fine mask} which aligns with the {\it coarse mask} (see \Figureref{teaser}).

The output from the transformer encoder layer is gradually scaled up from $40\times40$ to $320\times320$ in 3 steps. In each step, the image is first scaled up $2\times$ using bilinear interpolation and processed through a convolutional block described in Suppl.
Note that we also concatenate the appropriately resized input image to the input of each convolutional block. This information is pertinent for conditioning the {\it fine branch} to satisfy the regularization in \Subsectionref{gtv}. 

The features $\widehat{F}\in\R^{131\times320\times320}$ from the last convolutional block are linearly classified into $S_{\text{fine}}\in[0,1]^{320\times320}$ which is subsequently average pooled to $\widehat{S}_{\text{fine}}\in[0,1]^{40\times40}$ for aligning with the $S_{\text{coarse}}\in[0,1]^{40\times 40}$. The corresponding loss function is given as 
\begin{align}
    \calL_{\text{SR}}(X)\coloneqq\|\widehat{S}_{\text{fine}}-S_{\text{coarse}}\|_2^2.\label{sr}
\end{align}

\subsection{Graph total variation regularization (GTV)}
\label{subsec:gtv}

Graph-based regularization has yielded benefits in capturing high-frequency details of an image in \cite{guidedsuperresolution, pixtransform}. A similarity metric between pixels of an image $X$ is used to populate the affinity matrix $A>0$, which is then used to compute the degree matrix $D$. The graph Laplacian $L=D-A$ is used to compute the graph regularizer as the quadratic form for a graph signal~\cite{gsp} $s$, given by 
\begin{align}
    \calL_{reg}=\frac{1}{2}\sum_{(i,j)\in E}A_{ij}(s(i)-s(j))^2.\label{gtv}
\end{align}

Considering significant computational complexity from the total number of pairs of pixels, we enforce $A_{ij}=0$ when pixels $X_i$ and $X_j$ are not vertically or horizontally adjacent, also known as the pixel neighborhood $\mathcal{N}$.
This is equivalent to a weighted version of the total variation (TV) loss~\cite{tv2,tv3}, which has been previously used for denoising images and other signals~\cite{tv1,feat,tv3,coral}. A natural extension to graphs is discussed in \cite{gtv}.

\vspace{1mm}
\noindent{\bf GTV fine.} The guided super-resolution can result in more than one {\it fine mask} for a given {\it coarse mask}, which is where our graph total variation (GTV) loss not only works as a denoiser but plays a more important role as a regularizer. More specifically, $A_{ij}=\exp{\left(-\|X_i-X_j\|_2^2/\sigma\right)}$ is given by the euclidean similarity between the pairwise pixels. As a result, the $\calL_{\text{GTV-fine}}$ loss encourages the upsampler along the {\it fine} branch in \Figureref{sempart} to leverage the color information.

\vspace{1mm}
\noindent {\bf GTV coarse.} We also implement a similar graph TV regularizer denoted by $\calL_{\text{GTV-coarse}}$ for the {\it coarse mask} based on $A_{ij}=W_{ij}\mathbf{1}\{i\in\mathcal{N}(j)\}$ where $W_{ij}$ is as defined in \eqref{ncutaff}. This is responsible for denoising and predicting a smooth {\it coarse mask}.
\begin{figure*}
\begin{center}
\scalebox{0.93}{
\includegraphics[width=1\textwidth]{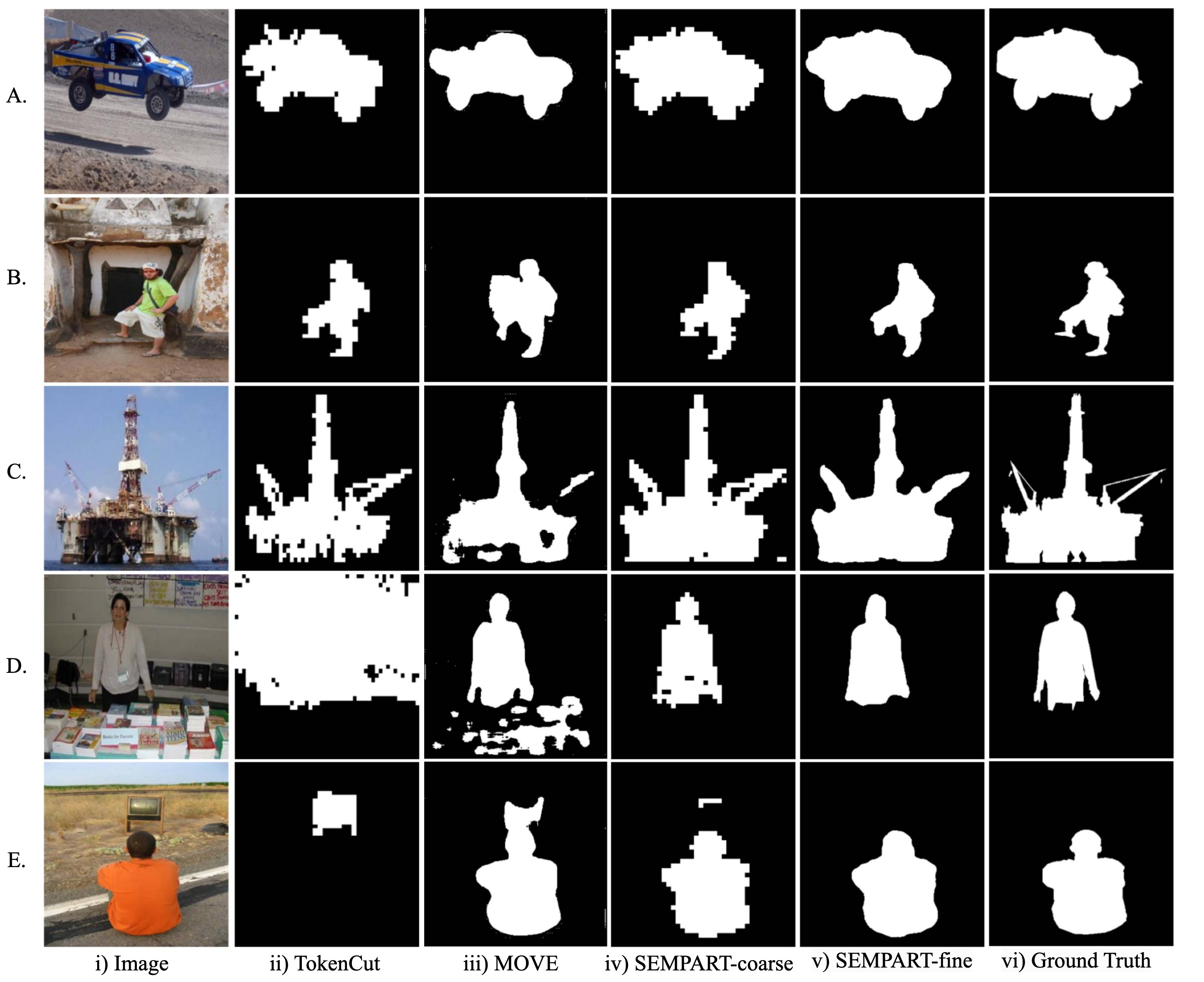}}
\end{center}\vspace{-0.5cm}
\caption{Qualitative comparison of \textsc{Sempart}-coarse and -fine with TokenCut~\cite{tokencut} and MOVE~\cite{move} for samples from DUT-OMRON~\cite{dutomron}. 
}\vspace{-0.6cm}
\label{fig:short}
\end{figure*}

\subsection{Loss formulation}
\label{subsec:loss}
The \textsc{Sempart} losses in \Subsectionref{sempart} together with the GTV losses in \Subsectionref{gtv} drive the joint learning of {\it coarse} and {\it fine masks}. While the \textsc{Sempart} losses are driven by DINO feature correspondences for inferring accurate image partitions, the GTV losses are significantly involved in denoising the predicted masks and regularizing the overall learning process. 
The loss functions for the {\it coarse} and {\it fine branches}, respectively, are, 
\begin{align}
\calL_{\text{coarse}}(x) &= \calL_{\text{Ncut}}(x)+\lambda_{\text{GTV-coarse}} \calL_{\text{GTV-coarse}}(x)\nonumber\\
\calL_{\text{fine}}(x) &= \lambda_{\text{GTV-fine}} \calL_{\text{GTV-fine}}(x) \nonumber\\
\calL_{\text{joint}}(x) &= \lambda_{\text{SR}} \calL_{\text{SR}}(x).
\end{align}    
This gives us our final expected self-supervised loss function $\calL_{\textsc{Sempart}}=\underset{x\sim\mathbb{P}(X)}{\mathbb{E}}[\calL_{\text{coarse}}(x)+\calL_{\text{fine}}(x) +\calL_{\text{joint}}(x)]$.

\section{Experiments}
\label{sec:expmt}
As done in \cite{tokencut,move}, we evaluate \textsc{Sempart} on unsupervised saliency segmentation and single object detection. 
\subsection{Implementation}
\label{subsec:implementation}
In our work, we use the self-supervised \cite{dino} ViT-s/8 transformer from the official implementation of DINO~\cite{dino}. DINO uses an $8\times8$ non-overlapping patch on a $3\times320\times320$ input and emits $384\times40\times40$ output which is provided to our simple transformer encoder layer and then routed through both the {\it coarse} and {\it fine branches} in \Figureref{sempart}. 
We employ Adam optimizer \cite{adam} with a learning rate of $0.0001$ and $\beta = (0.9, 0.999)$. We implemented \textsc{Sempart} in PyTorch and trained our models for 20 epochs with a batch size of 8 on a single NVIDIA Tesla P40 GPU.
Following careful consideration, hyperparameters
$\lambda_{\text{GTV-coarse}}=0.0006$, $\lambda_{\text{SR}}=20$, $\lambda_{\text{GTV-fine}}=0.0002$
have been applied for all results of \textsc{Sempart}.

\vspace{1mm}
\noindent{\bf Graph affinity.}
{\bf (a) Normalized cut.} Our implementation follows \cite{tokencut} in computing the affinity matrix $W$ based on \eqref{ncutaff} with a minor deviation. We set $W_{ii}=0$ to discard self-loops that do not belong to a graph cut. We show empirically that this improves model performance. Additionally we set $\tau=0.2$ and $\epsilon=$1e-6 in \eqref{ncutaff} for the $\calL_{\text{Ncut}}$ loss.
\noindent{\bf (b) GTV Coarse.} In addition to details provided in \Subsectionref{gtv}, we set $\tau=0$ and $\epsilon=$1e-6 for numerical stability.
\noindent{\bf (c) GTV Fine.} $\calL_{\text{GTV-fine}}$ regularizes the {\it fine mask} by limiting the possible solutions. The convolutional blocks learn to generate features that leverage both the contextual features from the transformer encoder and the RGB image features for predicting fine masks that mimic the {\it coarse mask} but also preserve the high-frequency image details.

In addition to details provided in \Subsectionref{gtv}, we also set $\sigma=1$. 

\vspace{1mm}
\noindent{\bf Foreground selection.}
We first binarize the indicator vector with threshold = $0.5$. 
In order to pick the foreground, we consider four strategies. {\bf (a)} Select the patch with a lower average distance to the center as the foreground. {\bf (b)} Discarding partitions with full spatial width or height as background, selecting the smaller partition to break a tie. {\bf (c)} Select the partition with greatest attention from the last layer of DINO. {\bf (d)} Select the partition occupying the least number of corners. If there is a tie, select the smaller partition. 

\subsection{Unsupervised saliency segmentation}
\label{subsec:saliency_segmentation}

\renewcommand{\arraystretch}{1.2}
\begin{table}
\begin{center}
\scalebox{0.79}{
\begin{tabular}{l|c|c|c|c|c}
\hline
\parbox{2cm}{\textbf{Method}} & \parbox{1.7cm}{\textbf{Avg. Time}} &\parbox{1.2cm}{\textbf{Model}} & \parbox{0.7cm}{\textbf{RES}} & \parbox{0.7cm}{\textbf{GPU}}& \parbox{0.7cm}{\textbf{CPU}} \\
\hline
TokenCut &  130ms &No &  Low &   Yes &Yes\\
TokenCut+BF &  337ms&No  &  High &   Yes&Yes \\
MOVE  &  13ms&  Yes&  High &  Yes&No\\
\textbf{\textsc{Sempart}}  &  14ms &Yes & High &  Yes&No \\
\hline
\end{tabular}}
\end{center}
\vspace{-0.3cm}
\caption{Both \textsc{Sempart} and MOVE train a model, generate high-resolution masks, and have comparable average inference times per image.}\vspace{-0.5cm}
\label{tab:timing}
\end{table}
\noindent {\bf Datasets.} As done in \cite{move,deepcut,selfmask}, we trained \textsc{Sempart} on the train split of DUTS~\cite{duts}, known as DUTS-TR and evaluate the performance of our model on the corresponding test split DUTS-TE~\cite{duts}, as well as DUT-OMRON~\cite{dutomron} and ECSSD~\cite{ecssd}. DUTS-TR contains 10,553 images, DUTS-TE contains $5,019$ images, DUT-OMRON contains 5,168 images, and ECSSD contains 1000 images. 

\vspace{1mm}
\noindent {\bf Evaluation.}
As done in \cite{move,tokencut}, we compute the per-pixel mask accuracy (Acc), intersection over union (IoU), and $\max F_\beta$~\cite{tokencut} for evaluating the performance of \textsc{Sempart}.
Accuracy is the fraction of pixels correctly predicted into the foreground or background. The overlap between the binary saliency mask and the ground truth gives IoU. We set $\beta=0.3$ as per \cite{move,tokencut} where $\max F_\beta$ is given for the threshold used for binarizing the mask that maximizes $F_\beta$. 

\vspace{1mm}
\noindent {\bf Results.}
We compared the performance of \textsc{Sempart} with recent state-of-the-art MOVE~\cite{move} and several other standard baselines referenced therein. \Tableref{main_results} contains three horizontal sections corresponding to the baseline method, followed by applying a bilateral filtering~\cite{bilateralfiltering} step. The final section involves generating pseudo ground truth based on the baseline method and training MaskFormer~\cite{maskformer} in a class agnostic manner, as in \cite{selfmask}.

We observe that applying the bilateral filter after inferencing \textsc{Sempart} on a per-image basis is detrimental to the overall performance, as is also seen in \cite{move}, with the performance of \textsc{Sempart}-Fine deteriorating significantly.

\textsc{Sempart} significantly outperformed all other baselines in all three sections across all datasets. Although \textsc{Sempart} is primarily motivated by the normalized cut minimization in \cite{tokencut}, the expected normalized cut loss in \Subsectionref{loss} co-optimized with the image-guided graph-based super-resolution loss results in significant improvement in performance. As seen in \Figureref{short}, the per-image optimization in TokenCut selects regions that are not salient or present in the foreground.
\begin{figure}[t]
\begin{center}
\includegraphics[width=\linewidth]{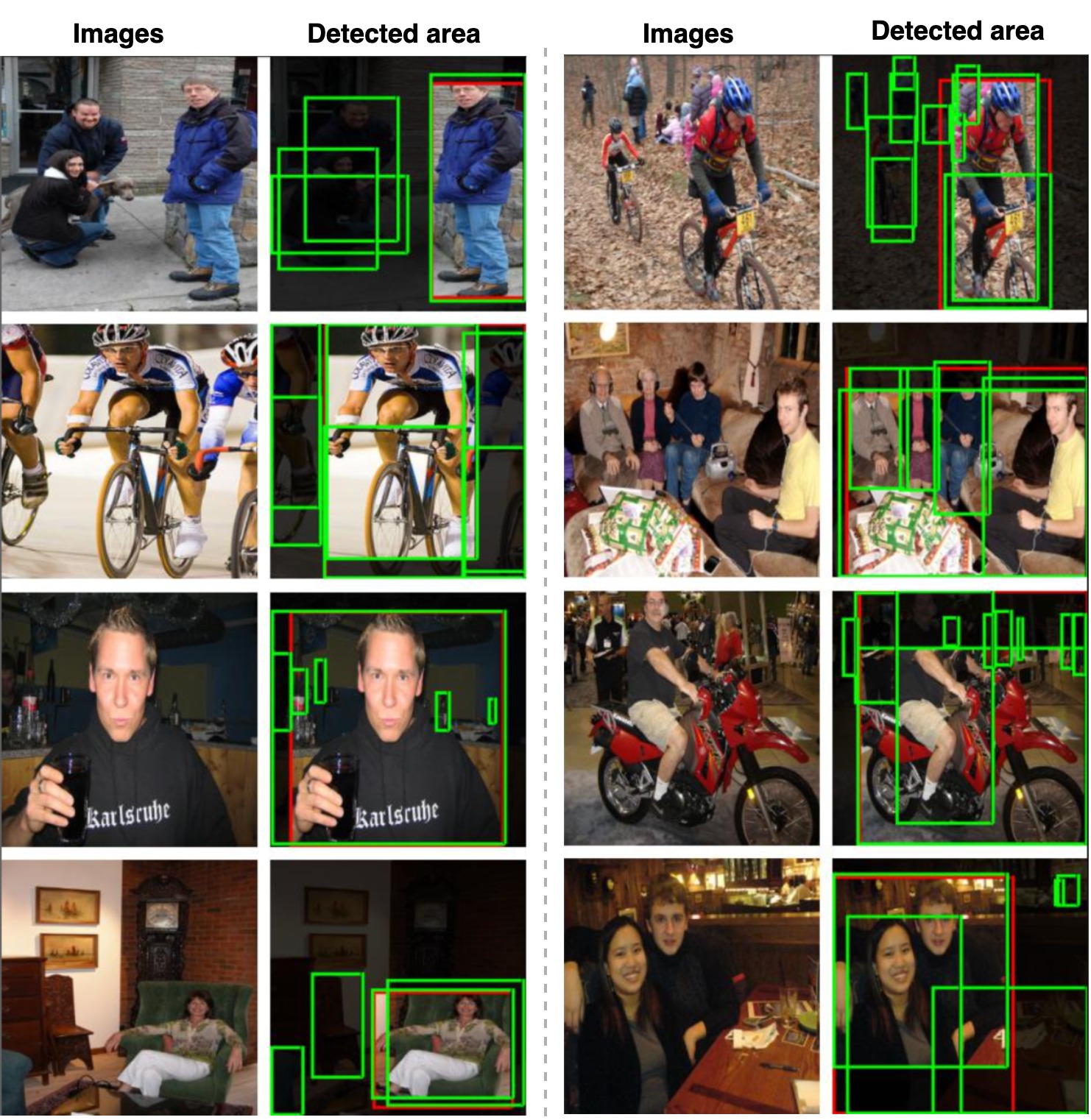}
\end{center}
\vspace{-0.5cm}
   \caption{\textsc{Sempart} for single object detection. Green boxes are ground truth bounding boxes, and the red box is our predicted bounding box. Intersection area is highlighted. }\vspace{-0.3cm}
\label{fig:long}
\label{fig:onecol}
\end{figure}

\textsc{Sempart} significantly outperforms the movability~\cite{move} heuristic in all three sections for all datasets. From \Figureref{short}, we find that MOVE may include multiple semantically unrelated patches into the movable object mask. Additionally, we note that MOVE greatly relies on retraining according to \textsc{SelfMask}~\cite{selfmask} for outperforming previous state-of-the-art. While \textsc{Sempart}-Coarse predicts noisy masks (see \Figureref{short}-A,C,D, and E) with slight errors as seen in the last example, \textsc{Sempart}-Fine results in refinement with improved ground truth alignment.

\subsection{Single object detection}
\label{subsec:single_object_detection}
\noindent {\bf Datasets.}
We evaluate our model on three datasets - the train split of COCO20K~\cite{cocodataset} and the training and validation splits of VOC07~\cite{voc07} and VOC12~\cite{voc12}. Each image in these datasets has one or more bounding boxes corresponding to each object. The objective is to localize any single object. 

\vspace{1mm}
\noindent {\bf Evaluation.}
We detect connected components for separating multiple objects for an image's \textsc{Sempart} mask\footnote{If multiple objects lie in a component this evaluation is less reliable.}. The component with the largest bounding box is used as the object prediction. Suppose the highest IoU between our predicted bounding box and all ground truth bounding boxes exceeds $0.5$. In that case, we treat it as a successful prediction and use this to compute \textit{Correct Localization} (CorLoc) metric which is simply the accuracy of prediction. 

\vspace{1mm}
\noindent {\bf Results.}
\textsc{Sempart} results in superior bounding-boxes which perform comparably with state-of-the-art MOVE, outperforming it on COCO20k dataset (see \Tableref{singleobject}). Our findings suggest that increasing $\tau$ to $0.25$ helps us prevent co-located disparate objects from lying in the same connected component and results in a slight improvement.

\renewcommand{\arraystretch}{1.2}
\begin{table}
\begin{center}
\scalebox{0.85}{
\begin{tabular}{l|c|c|c}
\hline
\parbox{2cm}{\textbf{Method}} & \parbox{1cm}{\textbf{VOC07}} & \parbox{1cm}{\textbf{VOC12}} & \parbox{1.5cm}{\textbf{COCO20K}} \\
\hline
DDT+~\cite{ddt}   &    50.2   &  53.1 &   38.2      \\
rOSD~\cite{rosd}   &   54.5   &   55.3    &   48.5      \\
LOD~\cite{lod}  &   53.6   &   55.1    &   48.5     \\
FreeSOLO~\cite{freesolo}  &   56.1  &   56.7    &   52.8     \\
LOST~\cite{lost}  &  61.9  &  64.0  &  50.7     \\
Deep Spectral~\cite{deepspectralclustering}  &  62.7  &  66.4  &   52.2 \\
TokenCut~\cite{tokencut} &  68.8  &  72.1 &   58.8 \\
MOVE~\cite{move}  &  \textbf{76.0}  &  \textbf{78.8} &  66.6 \\
\textbf{\textsc{Sempart}-Coarse}  &  74.7  & 77.4 &  \textbf{66.9} \\
\textbf{\textsc{Sempart}-Fine}  &  75.1 &  76.8 &  66.4 \\
\hline
\end{tabular}}
\end{center}
\vspace{-0.3cm}
\caption{\textsc{Sempart} bounding boxes exhibits a high CorLoc comparable to state-of-the-art MOVE~\cite{move}, for single object discovery on VOC2007 \cite{voc07}, VOC2012\cite{voc12} and outperforms it on COCO20K \cite{cocodataset} dataset.}
\label{tab:singleobject}\vspace{-0.4cm}
\end{table}
\begin{table}
\begin{center}
\scalebox{0.85}{
\begin{tabular}{l|c|c| c}
\hline
$\textbf{\parbox{3cm}{Method}}$ & $\textbf{OMRON*}$ & $\textbf{D-TE*}$ & $\textbf{ECSSD}$ \\
\hline
\textit{fs}: framing prior & 0.663 & 0.730 & 0.825 \\
\textit{fs}: centrality & 0.652 & 0.736 & 0.854 \\
\textit{fs}: total attention & {\bf 0.668} & 0.745 & 0.853  \\
\hline
w/ self-loops in $W$ & 0.667 & 0.743 & 0.846\\
\hline
w/o GTV coarse & 0.646& {\bf 0.749}& 0.848\\
w/o GTV fine &  0.637& 0.717& 0.818\\
\hline
train fine mask directly & 0.645 & 0.738 & 0.845 \\
\hline
w/o joint training & 0.662 & 0.743 & 0.849\\
\hline
\textbf{\textsc{Sempart}-Fine} & {\bf 0.668} & {\bf 0.749} & {\bf 0.855} \\
\hline
\end{tabular}}
\end{center}\vspace{-0.3cm}
\caption{Ablations of \textsc{Sempart} for saliency, using mIoU. *Shorthand has been used due to space constraints; OMRON refers to DUT-OMRON \cite{dutomron} and D-TE refers to DUTS-TE \cite{duts}; \textit{fs} denotes foreground selection. }
\label{tab:ablations}\vspace{-0.3cm}
\end{table}

\begin{figure}[ht]
\begin{center}
\scalebox{0.95}{
\includegraphics[width=1.0\linewidth]{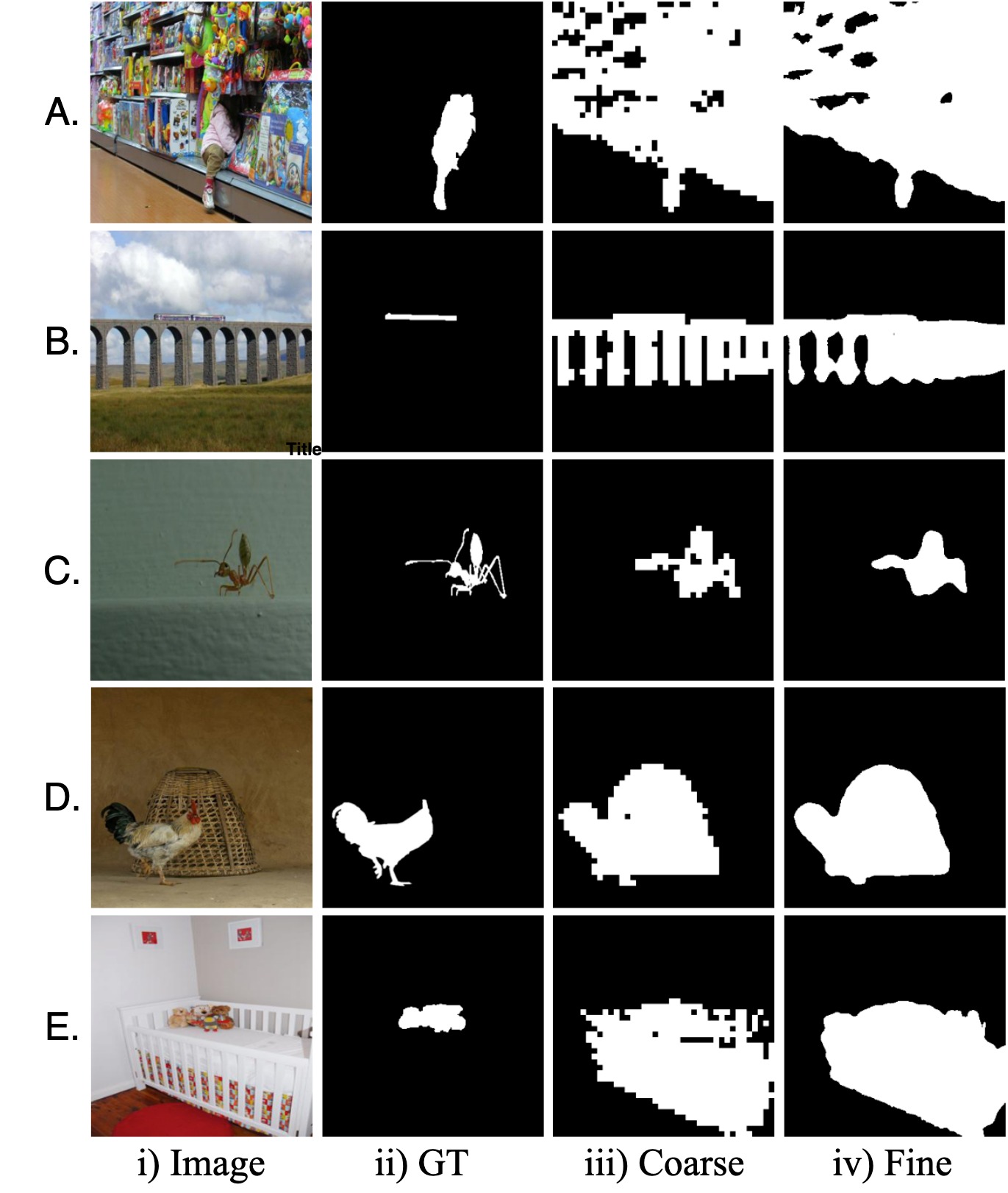}}
\end{center}
\vspace{-0.4cm}
\caption{Limitations of \textsc{Sempart}. Human bias towards humans and moving objects are shown in A and B. \textsc{Sempart} cannot capture the intricate details and smooths over narrow regions in B and C. An immovable background object is included, which is not as visually salient as the rooster in D.  The crib is the same color as the wall in E; therefore, the toys are prominent. However, DINO highlights the semantic differences for partitioning the entire crib from the background.}\vspace{-0.4cm}
\label{fig:limit}
\end{figure}
\subsection{Ablations}
\label{subsec:ablations}
\noindent We ablated \textsc{Sempart} for saliency segmentation as follows,

\vspace{1mm}
    \noindent \textbf{Foreground selection.} 
    Unlike \cite{move}, where the foreground is given by the {\it movable} object, \textsc{Sempart} selects partitions based on occupying {\it least corners} given by \textsc{Sempart}-Fine in \Tableref{ablations}. Motivated by \cite{selfmask,lost}, we compare with selection based on closeness to the image center ({\it centrality}), as well as the {\it framing prior}~\cite{selfmask}, which labels the segment occupying full spatial width or height as background while breaking ties based on selecting the smaller partition as foreground. Another heuristic that is a close contender to {\it least corners} is {\it total attention}, in which the partition having the highest total overlap with the DINO \texttt{[CLS]} token attention map as foreground.

\vspace{1mm}
    \noindent \textbf{Self-loops.} We populate $W_{ii}$ with \eqref{ncutaff} instead of 0 for $\calL_{\text{Ncut}}$ and demonstrate that the performance deteriorates.
    
\vspace{1mm}
    \noindent \textbf{Graph TV regularization.} 
    \textsc{Sempart} without either $\calL_{\text{GTV-coarse}}$ or $\calL_{\text{GTV-fine}}$ is detrimental to performance. The absence of the GTV-fine loss has a greater negative impact.

    \vspace{1mm}
    \noindent \textbf{Training fine mask directly.} We evaluate a setting where we only have a {\it fine branch} (see \Figureref{sempart}), and the $\calL_{\text{Ncut}}$ and $\calL_{\text{GTV-fine}}$ losses. \Tableref{ablations} demonstrates that this is inferior to \textsc{Sempart} despite being almost equivalent in the number of parameters. We attribute this to the absence of the {\it coarse branch} and the corresponding $\calL_{\text{Ncut}}$ loss which in turn regularized the transformer encoder for subsequent consumption by the convolutional blocks.

    \vspace{1mm}
    \noindent \textbf{Joint training.} We evaluate a variant of \textsc{Sempart}, where the {\it coarse} and {\it fine branch} are trained independently. While the $\calL_{\text{coarse}}$ only optimizes the {\it coarse branch} and the transformer encoder (see deviations from \Figureref{sempart} in Suppl.), the gradients from $\calL_{\text{fine}}$ and $\calL_{\text{joint}}$ are prohibited from optimizing these modules.
    As seen in \Tableref{ablations}, this is detrimental to performance on all datasets, verifying our hypothesis that co-optimizing {\it coarse} and {\it fine} mask is mutually beneficial.

\subsection{Limitations}
\label{subsec:limitations}
Visual saliency is not agnostic to various human biases in favor of humans and animals, as well as objects which are likely to move in a subsequent frame or have high contrast with the background. \Figureref{limit} discusses examples where the ground truth favors a human, a train crossing a bridge, and a rooster over all other objects. \textsc{Sempart} results in over-selection here as it does not explicitly incorporate these priors or even control the object size. Furthermore, our graph TV loss can sometimes merge narrow co-located regions into the mask, as seen in the \Figureref{limit}-B and C, which can also be detrimental to localizing objects.

\section{Conclusion}
\label{sec:conclusion}
\textsc{Sempart} demonstrates the efficacy of graph-driven objectives towards self-supervised image partitioning and establishes state-of-the-art performance for detecting salient regions and a competitive advantage in localizing objects. We address the limitations of expensive post-processing, limited resolution, and noise artifacts in saliency masks. We demonstrate the value of a joint learning paradigm for inferring high-quality masks at multiple resolutions using \textsc{Sempart}, which will hopefully be a vital enabler of the subsequent investigation into class-aware object detection for diverse vision systems.\vspace{3mm}

\noindent{\bf Acknowledgements} The authors gratefully thank Ambareesh Revanur and Deepak Pai for their valuable feedback, and the anonymous reviewers for their comments.

{\small
\bibliographystyle{ieee_fullname}
\bibliography{egbib}
}

\newpage 
\clearpage
\appendix

\section{Notation}
\label{app:notation}
\begin{figure*}[hbt!]
\begin{center}
\scalebox{0.98}{
\includegraphics[width=1\textwidth]{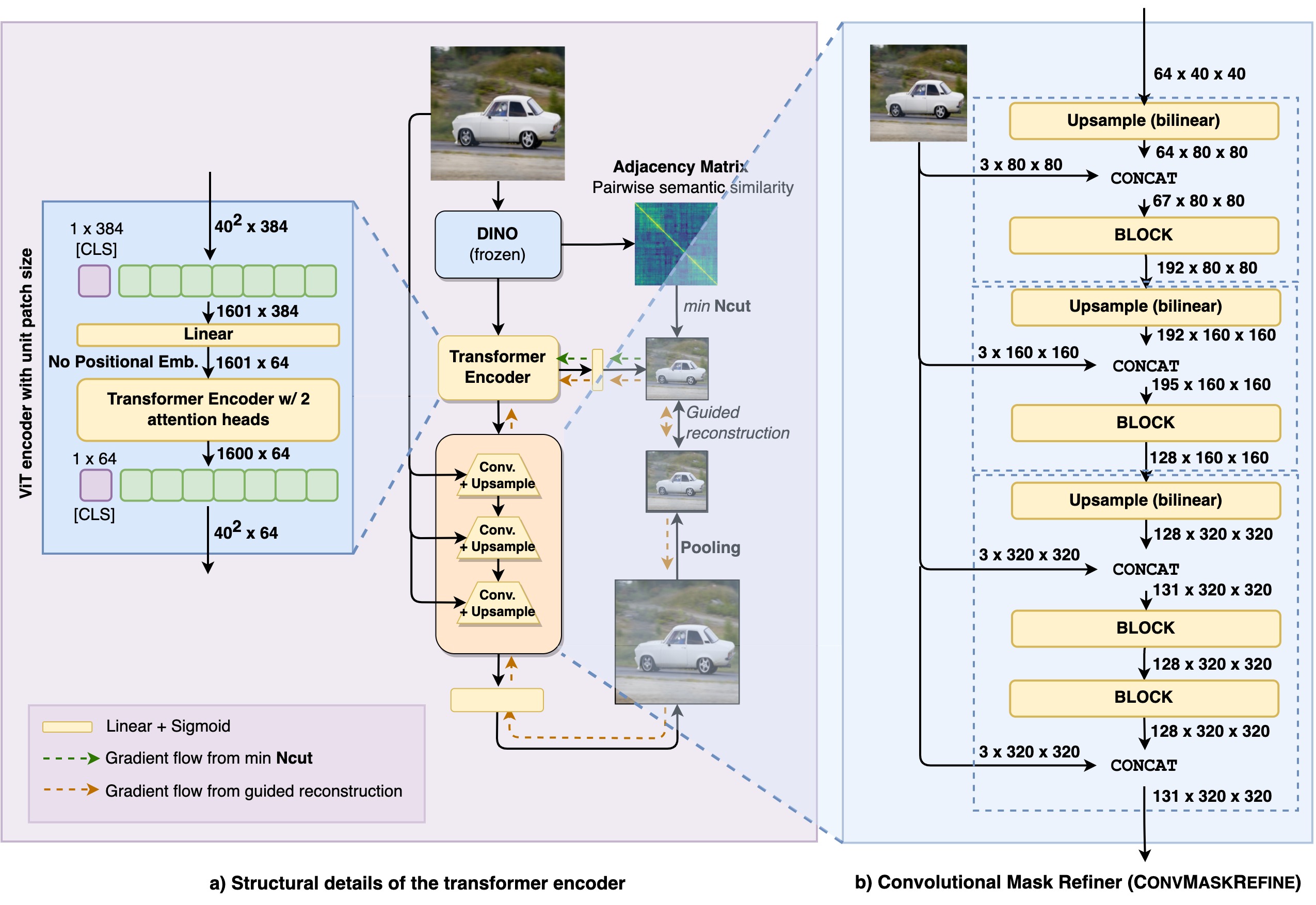}}
\end{center}
\caption{Expanded overview of \textsc{Sempart}: In addition to the details presented in \Figureref{sempart}, we zoom in to the transformer encoder in \Figureref{sempartarch} (a) and the convolutional mask refinement network in \Figureref{sempartarch} (b). \textsc{Block} is as defined in \eqref{block}.}\vspace{-0.5cm}

\label{fig:sempartarch}
\end{figure*}
For an image $X\in\R^{3\times 320\times 320}$, we represent the self-supervised features of DINO~\cite{dino} obtained for all $8\times 8$ non overlapping patches as $F\in\R^{384\times 40\times 40}$. We use $\calL$ to denote loss functions. $\mathbb{P}(M)$ and $\mathbb{E}[M]$ denote the distribution and the expected value of the random variable $M$. $\mathbf{1}\{\cdot\}$ is used to denote the indicator function.

For a graph, $G=(V, E)$, $V$, and $E$ denote the vertex and edge set, respectively. $W$ and $A$ represent the adjacency or affinity matrix for the $\calL_{\text{Ncut}}$ in \Subsectionref{sempart} and the GTV losses in \Subsectionref{gtv} respectively.

$I$ denotes the identity matrix. $D$ and $L$ correspond to the degree matrix and the Laplacian matrix for the graph $G$, respectively. $s:v\in V\rightarrow s(v)\in R$ has been used to denote a scalar signal as a function defined over the graph's nodes $v\in V$ as the domain. The definition of $S$ naturally follows as $S\coloneqq[s(1),s(2),\ldots,s(|V|)]^T$. 

\section{Architecture for \Sectionref{approach}}
\label{app:architecture}
\Sectionref{approach} describes the essential details of the \textsc{Sempart} architecture in which we emphasize the importance of two vital learnable components: {\bf (a)} the transformer encoder as a shared parametrized module between both the {\it coarse} and {\it fine branch}, {\bf (b)} the convolutional mask refinement network for generating high resolution {\it fine masks}. \Figureref{sempartarch} (a) and (b) presents the transformer encoder as well as the convolutional mask refinement network, respectively, in greater detail. Furthermore, we also elaborate upon these individual modules in \Appendixref{pseudo}.

\section{Pseudocode for \Sectionref{approach}}
\label{app:pseudo}
\textsc{Sempart} is a self-supervised multi-resolution image bi-partitioning heuristic that successfully distills the encoded information from DINO~\cite{dino} towards high-quality unsupervised semantically meaningful partitions that significantly resonate with the notion of visual saliency for an image. In this section, we elaborate upon the forward pass described in \Subsectionref{sempart} to \Subsectionref{loss} culminating in \Algorithmref{sempart}.

\vspace{1mm}
\noindent {\bf DINO backbone~\cite{dino}:} DINO~\cite{dino} is a widely adopted self-supervised vision model which emits features that are contextually aware and captures the semantic richness of an image (see~\cite[Figure 1]{dino}). \textsc{Sempart} leverages the self-supervised \cite{dino} ViT-s/8 transformer based on \cite{vit} from the official implementation of DINO~\cite{dino}, which processes a $320\times 320$ image $X$ as a $40\times 40$ positionally aware flattened sequence of $8\times 8$ non overlapping patches. We denote the transformation by \begin{align}
    \textsc{DINO}(X):X\in \R^{3\times 320\times 320}\rightarrow F\in \R^{384\times 40\times 40}.
\end{align}
Note that in fact DINO emits $\R^{384\times(1+  40\times 40)}$, however we discard the \texttt{[CLS]} token feature for subsequent modules. In our implementation, the DINO backbone remains frozen.

\vspace{1mm}
\noindent {\bf Transformer encoder~\cite{vit}:} We apply a single layer transformer encoder\footnote{Implementation is borrowed from \cite{vit}.} with two attention heads that transform $F\in \R^{384\times 40\times 40}$ to $\widetilde{F}\in\R^{64\times 40\times 40}$.\begin{align}
\widetilde{F}\leftarrow\textsc{TransformerEncoder}(F).\label{trfenc}
\end{align}
Emitted features $\widetilde{F}$ are shared between both the \textsc{Sempart}-Coarse and \textsc{Sempart}-Fine branches (see \Figureref{sempart}).

\vspace{1mm}
\noindent {\bf Convolutional mask refinement network (\Subsectionref{sempart}):} As also done in \cite{move}, we define ${\textsc{Block}_{\text{out\_ch}}^{\text{in\_ch}}}$ as 
\begin{align}
    3\times 3 {\textsc{ Conv}_{\text{out\_ch}}^{\text{in\_ch}}}\rightarrow \textsc{BatchNorm}\rightarrow\textsc{LeakyReLU}\label{block}
\end{align}
where $K\times K {\textsc{ Conv}_{\text{out\_ch}}^{\text{in\_ch}}}$ is a padded $K\times K$ convolution with stride = 1, $\text{in}\_\text{ch}$ and $\text{out}\_\text{ch}$ correspond to the number of input and output channels respectively. Before each block, we also concatenate - denoted by the $||_c$ operator - an appropriately resized image along the channel dimension. 

Consequently, our convolutional mask refinement network is given by alternating bilinear \textsc{Upsample} and \textsc{Block} as follows
\begin{align}
    \widetilde{F}'&\leftarrow{\textsc{Block}_{192}^{67}}\left[{\textsc{Upsample}_{\text{bilinear}}^{2\times 2}}\left(\widetilde{F}\right) ||_c X^{3\times80\times80}\right]\nonumber\\
    \widetilde{F}''&\leftarrow{\textsc{Block}_{128}^{195}}\left[{\textsc{Upsample}_{\text{bilinear}}^{2\times 2}}\left(\widetilde{F}'\right) ||_c X^{3\times160\times160}\right]\nonumber\\
    \widetilde{F}'''&\leftarrow{\textsc{Block}_{128}^{131}}\left[{\textsc{Upsample}_{\text{bilinear}}^{2\times 2}}\left(\widetilde{F}''\right) ||_c X^{3\times320\times320}\right]\nonumber\\
    \widehat{F}&\leftarrow{\textsc{Block}_{128}^{128}}\left(\widetilde{F}'''\right)||_c X^{3\times320\times320}.\label{convref1}
\end{align}
The image $X$ is provided as side information and is essential for conditioning the convolutional mask refinement network towards generating {\it fine masks} driven by the $\calL_{\text{GTV-fine}}$ loss. We modularize the complete convolutional mask refinement transformation given in \eqref{convref1} as follows,
\begin{align}
    \widehat{F}\leftarrow\textsc{ConvMaskRefine}(\widetilde{F},X).\label{convref2}
\end{align}

\vspace{1mm}
\noindent {\bf Coarse branch (\Subsectionref{sempart}):} The {\it coarse branch} applies a binary linear classification head (\textsc{LCH}) as a composition of a linear layer followed by sigmoid to $\widetilde{F}$, resulting in $S_{\text{coarse}}\in[0,1]^{40\times 40}$. 
\begin{align}
    S_{\text{coarse}}\leftarrow \textsc{LCH}_1^{64}\left(\widetilde{F}\right).\label{bch1}
\end{align}
Here $\textsc{LCH}_1^{\text{in}\_\text{ch}}$ corresponds to
\begin{align}
    1\times 1\textsc{ Conv}_{1}^{\text{in}\_\text{ch}}\rightarrow \textsc{sigmoid}.
\end{align}
We denote this operation as follows
\begin{align}
    S_\text{coarse}\leftarrow\textsc{CoarseBranch}(\widetilde{F}).
\end{align}

\vspace{1mm}
\noindent {\bf Fine branch (\Subsectionref{sempart}):} The fine branch involves the composition of the \textsc{TransformerEncoder} features $\widetilde{F}$ with convolutional mask refinement network in \eqref{convref2}, which produces $\widehat{F}$. Along the lines of \eqref{bch1}, a binary classification head is subsequently applied as follows
\begin{align}
    S_{\text{fine}}\leftarrow \textsc{LCH}_1^{131}\left(\widehat{F}\right)
\end{align}
Here $S_{\text{fine}}\in[0,1]^{320\times 320}$ is the high resolution {\it fine mask}. Therefore we denote the {\it fine branch} as 
\begin{align}
    S_\text{fine}\leftarrow\textsc{FineBranch}(X, \widetilde{F}).
\end{align}
where \textsc{FineBranch} is given by 
\begin{align}
    \textsc{ConvMaskRefine}\rightarrow \textsc{LCH}_1^{131}
\end{align}

\vspace{1mm}
\noindent {\bf \textsc{Sempart} (\Subsectionref{loss}):} The loss functions described in \Subsectionref{sempart} and \Subsectionref{gtv} are motivated by graph-based bi-partitioning of images based on deep semantic correspondences between regions as well as driven by graph total variation of the generated masks over the entire image. This results in high-quality self-supervised masks based on principles of normalized cut and guided super-resolution. We compute the corresponding loss functions in \Subsectionref{loss} to give us the eventual \textsc{Sempart} loss in \Algorithmref{sempart}.
\begin{algorithm}[h]
  \caption{\textsc{Sempart}}
  \label{algo:sempart}
  \hspace*{\algorithmicindent} \textbf{Input} $X\in \R^{3\times320\times 320}$ in RGB space\\
  \hspace*{\algorithmicindent} \textbf{Output} Loss $\mathcal{L}_{\textsc{Sempart}}$
  \begin{algorithmic}[1]
  \Function{Loss}{$X$}
  \State $F= \textsc{DIN
  O}(X)$
  \State $\widetilde{F}=\textsc{TransformerEncoder}(F)$
  \State $S_\text{coarse}=\textsc{CoarseBranch}(\widetilde F)$
  \State $S_\text{fine}=\textsc{FineBranch}(X, \widetilde F)$
  \State $\calL_{\text{Ncut}}=\calL_{\text{Ncut}}(F,S_\text{coarse})$\hfill See \eqref{ncut2}
  \State $\calL_{\text{GTV-coarse}}=\calL_{\text{GTV-coarse}}(F,S_\text{coarse})$\hfill See \Subsectionref{gtv}
  \State $\calL_{\text{SR}}=\calL_{\text{SR}}(S_\text{coarse},S_\text{fine})$\footnotemark\hfill See \eqref{sr}
  \State $\calL_{\text{GTV-fine}}=\calL_{\text{GTV-fine}}(X,S_\text{fine})$\hfill See \Subsectionref{gtv}
  \State $\calL_{\text{coarse}}=\calL_{\text{Ncut}}+\lambda_{\text{GTV-coarse}} \calL_{\text{GTV-coarse}}$\vspace{1mm}
  \State $\calL_{\text{fine}} = \lambda_{\text{GTV-fine}} \calL_{\text{GTV-fine}}$
  \State $\calL_{\text{joint}} = \lambda_{\text{SR}} \calL_{\text{SR}}$
  \State $\calL_{\textsc{Sempart}}=\calL_{\text{coarse}}+\calL_{\text{fine}} +\calL_{\text{joint}}$\hfill See \Subsectionref{loss}
  \State \Return $\calL_{\textsc{Sempart}}$
    \EndFunction
  \end{algorithmic}
\end{algorithm}
  \footnotetext{Note that this involves an average pooling step for aligning the spatial dimensions. See section on guided super-resolution in \Subsectionref{sempart}.}

The parameters of the transformer encoder, the convolutional mask refinement network, and the two binary classification heads are refined iteratively as per the loss $\calL_{\textsc{Sempart}}$. Note that this is an entirely unsupervised scheme where the DINO feature correspondences serve as the key source of self-supervision.

\section{Supplementary material for \Sectionref{expmt}}
\label{app:addres}
\begin{figure*}[hbt!]
\begin{center}
\scalebox{0.98}{
\includegraphics[width=1\textwidth]{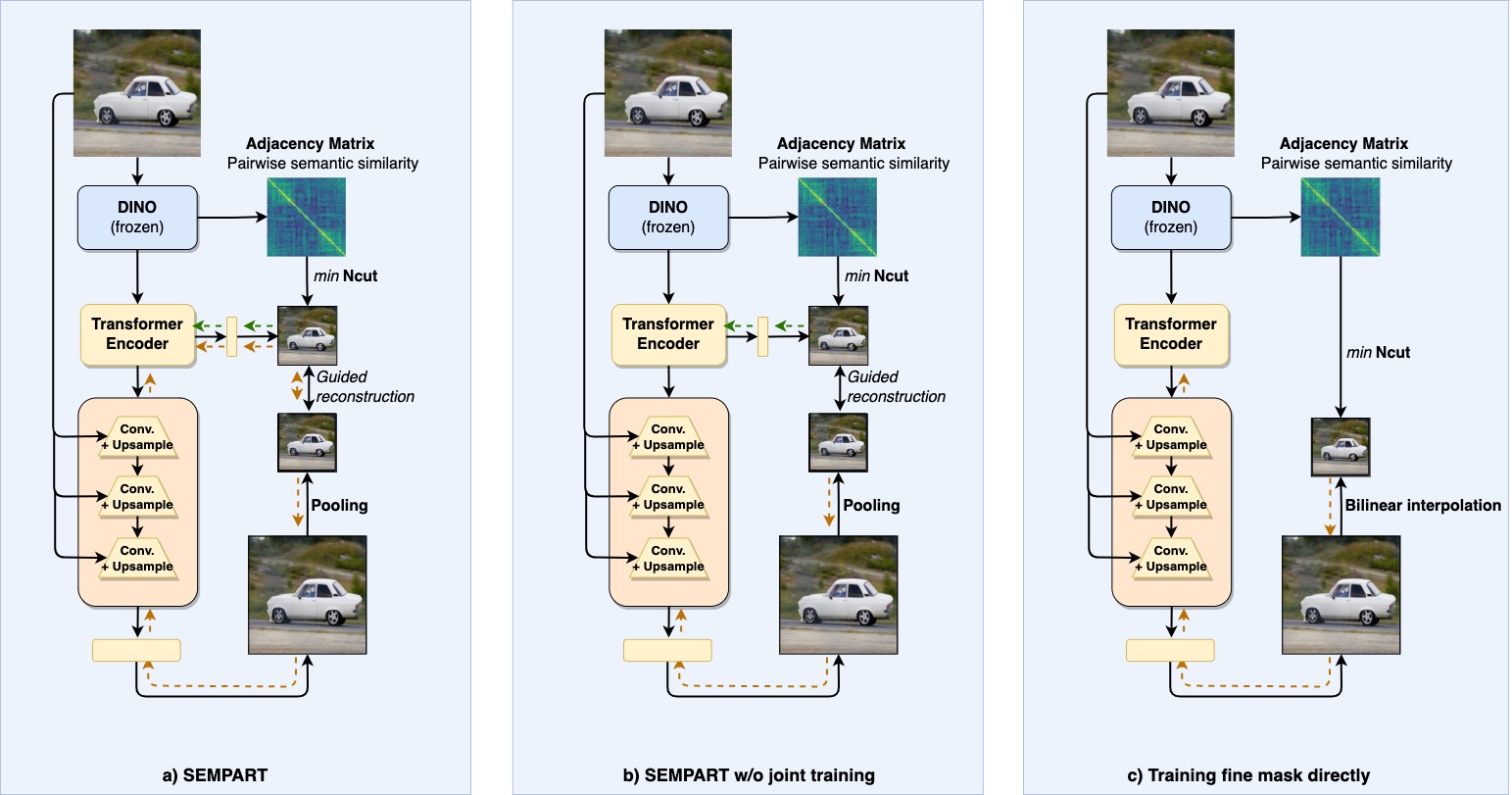}}
\end{center}

\caption{Comparison of \textsc{Sempart} with ablations of its architecture in decreasing order of performance from (a) to (c) (see \Tableref{ablations}).}

\label{fig:sempartcomparison}
\end{figure*}

\noindent{\bf Architecture ablation comparison.}
\Figureref{sempartcomparison} demonstrates the architectural differences between \textsc{Sempart}, and the ablations we compare with. In particular, as discussed in \Subsectionref{ablations}, we demonstrate the value of co-optimizing our {\it coarse} and {\it fine branches} (see \Figureref{sempartcomparison} (a)) as compared to only having the {\it fine branch} (see \Figureref{sempartcomparison} (c)) or having both branches trained independently (see \Figureref{sempartcomparison} (b)). Results of the paired Wilcoxon signed-rank test \cite{wilcoxon}  on the IoU metric, shown in \Tableref{ablations_pvalues}, confirm the value of architectural choices, using significance level of $0.05$.

\begin{table}[h]
\begin{center}
\scalebox{0.68}{
\begin{tabular}{l|c|c|c}
\hline
$\textbf{\parbox{3.5cm}{Method}}$ & $\textbf{DUT-OMRON}$ & $\textbf{DUTS-TE}$ & $\textbf{ECSSD}$ \\
\hline
 w/o GTV coarse & $0.646 \,(<{0.001})$& {\bf $\mathbf{0.749\, (-)}$}& $0.848\, (<{0.001})$\\
 w/o GTV fine &  $0.637\, (<{0.001})$& $0.717\, (<{0.001})$& $0.818\, (<{0.001})$\\
\hline
 train fine mask directly & $0.645\, (<{0.001})$ & $0.738\, (<{0.001})$ & $0.845\, (<{0.001})$ \\
\hline
w/o joint training & $0.662\, (<{0.001})$ & $0.743\, (<{0.001})$ & $0.849\, ({0.007})$\\
\hline
\textbf{\textsc{Sempart}-Fine} & {\bf $\mathbf{0.668}$} & {\bf $\mathbf{0.749} $} & {\bf $\mathbf{0.855}$} \\
\hline
\end{tabular}}
\end{center}\vspace{-0.5cm}
\caption{Ablations of \textsc{Sempart} for saliency, using mIoU~($p$-value). 
}
\label{tab:ablations_pvalues}\vspace{-0.4cm}
\end{table}

As described in \Subsectionref{ablations}, \Figureref{sempartcomparison} (b) demonstrates that normalized cut loss only affects the transformer encoder and the {\it coarse branch}. In contrast, the gradients from the guided reconstruction only affect the {\it fine branch}. The gradients from the corresponding GTV losses also only affect the respective branches. In \Figureref{sempartcomparison} (c), however, the {\it coarse branch} is completely discarded, and the fine branch is utilized both for optimizing the expected normalized cut loss as well as the corresponding $\calL_{\text{GTV-fine}}$ loss. 

In our experiments (see \Tableref{ablations}), we observe that the performance in terms of the mean IoU of unsupervised saliency detection deteriorates consistently across all our evaluation datasets as we go from \Figureref{sempartcomparison} (a) to (b) to (c). This aligns with our intuition by demonstrating that not only is there value in separately inferring a {\it coarse mask} using the {\it coarse branch}, which effectively has the impact of a regularizer of the \textsc{TransformerEncoder}, but it is also beneficial to co-optimize the {\it fine} branch with the {\it coarse branch}.

\begin{table}
\begin{center}
\scalebox{0.84}{
\begin{tabular}{l|c|c| c}
\hline
$\textbf{\parbox{3cm}{Method}}$ & $\textbf{OMRON*}$ & $\textbf{D-TE*}$ & $\textbf{ECSSD}$ \\
\hline
\textsc{Sempart}-Fine & 0.668 & 0.749 & 0.855 \\
\textsc{Sempart}-Fine\dag & 0.673 & 0.755 & 0.857 \\
\textsc{Selfmask} on \textsc{Sempart}-Fine & 0.698 & 0.749 & 0.850 \\
\hline
$\textsc{U}^2\textsc{-Net}$\cite{u2net} & 0.693 & 0.733 & 0.878 \\
\textsc{SelfReformer\cite{selfreformer}} & 0.744 & 0.830 & 0.900 \\
\hline
\end{tabular}}
\end{center}\vspace{-0.1cm}
\scriptsize\dag indicates that validation images were included during unsupervised training. \vspace{-0.3cm}
\caption{We compare \textsc{Sempart} variants with $\textsc{U}^2\textsc{-Net}$ and \textsc{SelfReformer} both of which are supervised.}\vspace{-0.1cm}
\label{tab:supervised}\vspace{-0.3cm}
\end{table}

\vspace{1mm}
\noindent {\bf Comparison with supervised methods.}
\Tableref{supervised} compares the performance of \textsc{Sempart} with recent state-of-the-art supervised methods~\cite{u2net,selfreformer}. 
We show that using \textsc{Sempart} masks for SelfMask training results in high quality masks outperforming the supervised $\textsc{U}^2\textsc{-Net}$ on DUT-OMRON and DUTS-TE. However, a more recent supervised method \cite{selfreformer} still outperforms \textsc{SEMPART} by a significant margin. 

We also observe that scaling the training set to also include the validation images improves the performance of \textsc{Sempart}, indicated by \textsc{Sempart}-Fine\dag.
\vspace{2mm}

\noindent {\bf Comparison with alternate backbones.}
Our experiments with alternate backbones in \Tableref{backbone}, indicates that the \textit{degree of pixelation} (DoP),  defined as the ratio of patch to image areas affects the performance. A larger ViT patch size is detrimental, and SSL features with lower DoP result in superior \textsc{Sempart} saliency masks (\Tableref{backbone} A,\,B vs.\,C,\,D). Nevertheless, the \textit{fine mask} always outperforms its accompanying \textit{coarse mask} by preserving high-frequency details.
\vspace{-5mm}
\begin{table}[h]
\begin{center}
\scalebox{0.62}{
\begin{tabular}{c|c|c|c|c|c|c|c|c}
\hline
&{\bf Backbone} & \textbf{Arch} & \textbf{Type}& {\bf Input} & {\bf DoP} & {\bf OMRON} & {\bf D-TE} &{\bf ECSSD} \\
\hline
{ A.}& {\multirow{4}{*}{\shortstack{DINOv2\\(2023)}}} & {\multirow{4}{*}{ViT-S/14}}  &{Coarse} & {$224^2$} & {$3.9\textrm{e-}3$} & {0.460} & {0.539} & {0.659}\\
{B.}& & &{Fine}   & {$224^2$} & {$3.9\textrm{e-}3$} & {0.523} & {0.598} & {0.717}\\

\cline{4-9}
{ C.}&  & &{Coarse}   & {$560^2$} & {$6.25\textrm{e-}4$} & {$0.554$} & {$0.671$} & {$0.773$}\\
{D.}&  & &{Fine}   & {$560^2$} & {$6.25\textrm{e-}4$} & {$0.57$} & {$0.686$} & {$0.796$}\\
\hline
{E.}& \multirow{4}{*}{DINO} & {\multirow{2}{*}{ViT-S/16}} & {Coarse} & {$320^2$} & {$2.5\textrm{e-}3$} & {0.573} & {0.640} & {0.766}\\
{F.}& & &{Fine} & {$320^2$} & {$2.5\textrm{e-}3$} & {0.596} & {0.656} & {0.793}\\

\cline{3-9}
G.&  &\multirow{2}{*}{ViT-S/8} & Coarse & $320^2$ & $6.25\textrm{e-}4$ & 0.640 & 0.727 & 0.837\\
H.&  & &Fine & $320^2$ & $6.25\textrm{e-}4$ & {\bf 0.668} & {\bf 0.749} & {\bf 0.855}\\
\hline
\end{tabular}}
\end{center}
\vspace{-0.3cm}

\caption{\textsc{Sempart} IoU (last three columns) for DINOv2 and DINO.}
\vspace{-0.4cm}
\label{tab:backbone}
\end{table}

\begin{figure}[ht]
\begin{center}
\scalebox{0.98}{
\includegraphics[width=1.0\linewidth]{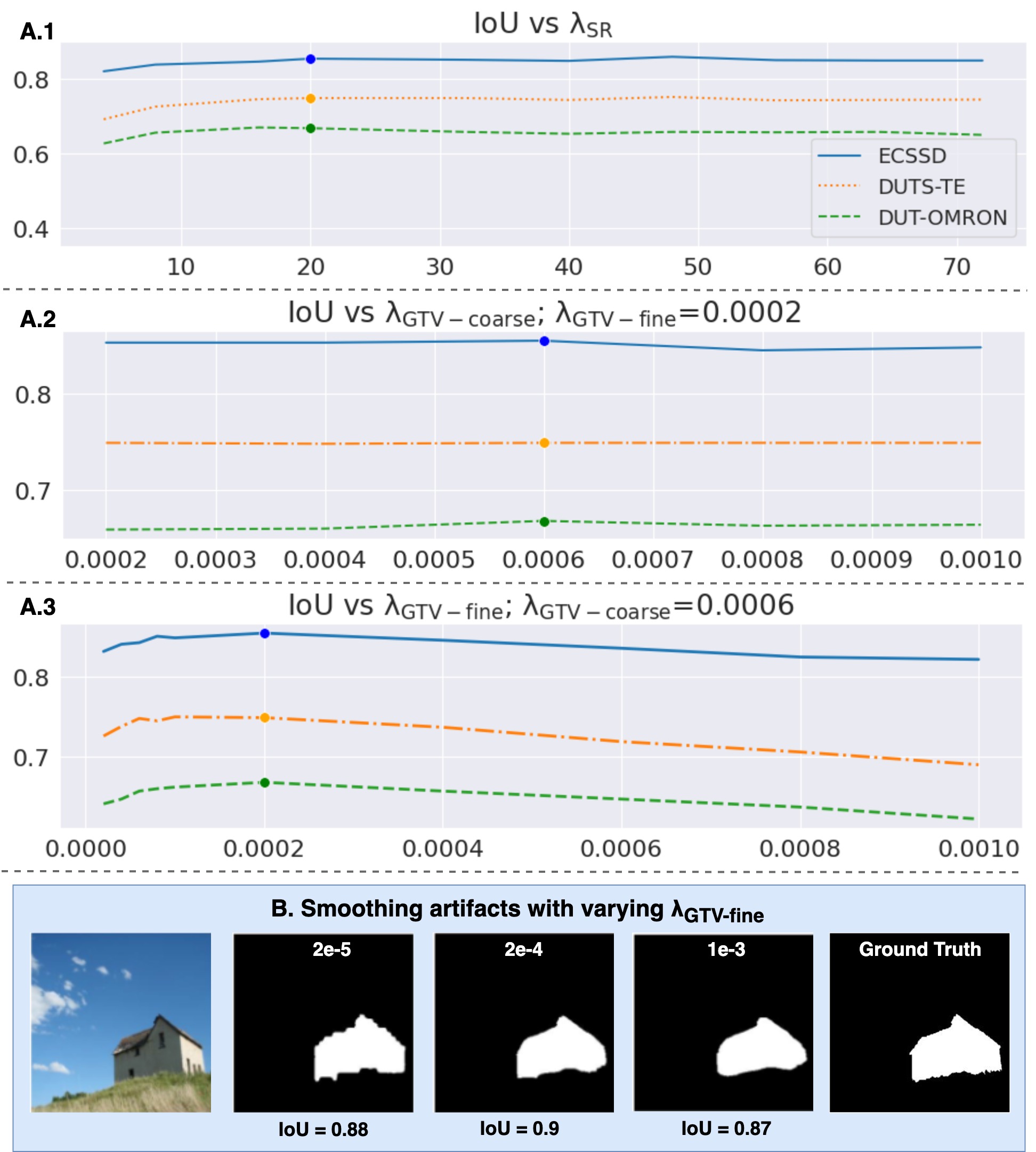}}
\end{center}
\vspace{-0.4cm}
\caption{Hyperparameter sensitivity analysis of \textsc{SEMPART}-Fine.}\vspace{-0.4cm}
\label{fig:hyp_sens}
\end{figure}

\noindent {\bf Hyperparameter sensitivity analysis.} \Figureref{hyp_sens} (A.1, A.2) show that the performance is typically robust to changes in $\lambda_{\text{SR}}$ and $\lambda_{\text{GTV-coarse}}$ respectively. \Figureref{hyp_sens} (A.3, B) show that the performance suffers with low and high $\lambda_{\text{GTV-fine}}$ values due to jaggedness and over-smoothing respectively.

\vspace{2mm}

\begin{figure*}[hbt!]
\begin{center}
\scalebox{0.95}{
\includegraphics[width=1\textwidth]{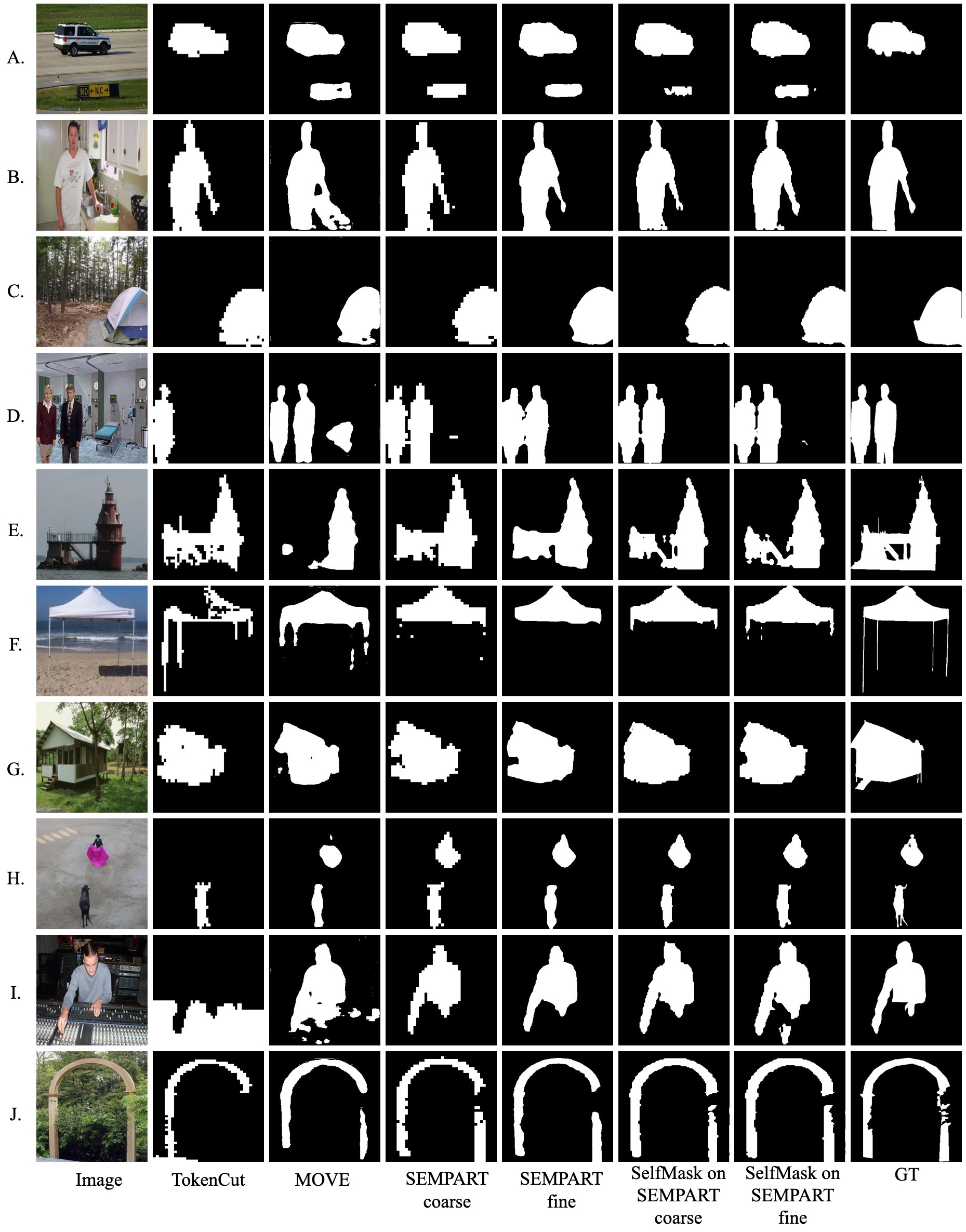}}
\end{center}
\caption{Additional examples on the DUT-OMRON~\cite{dutomron} dataset.}
\label{fig:add1}
\end{figure*}

\begin{figure*}[hbt!]
\begin{center}
\scalebox{0.95}{
\includegraphics[width=1\textwidth]{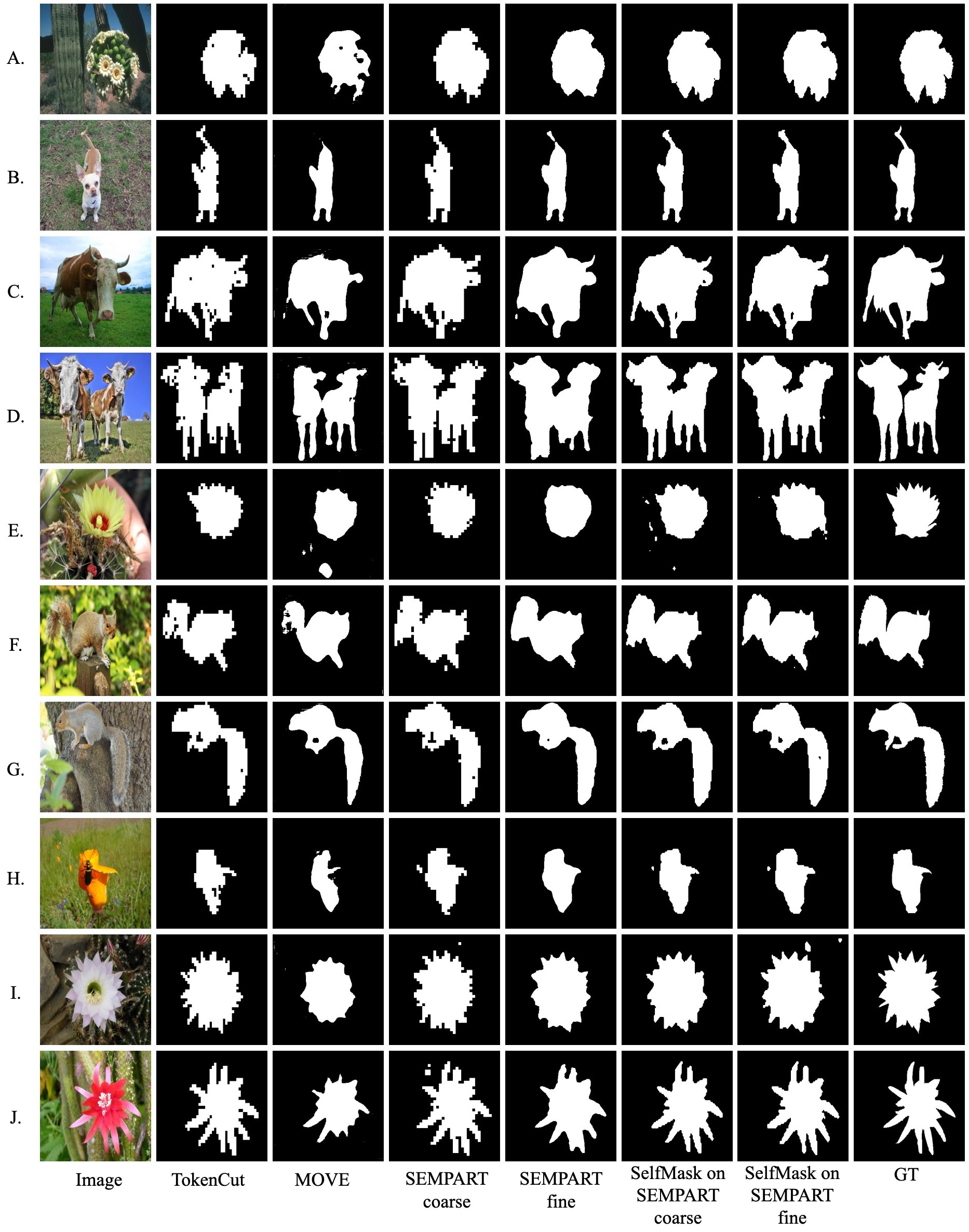}}
\end{center}
\caption{Additional examples on the ECSSD~\cite{ecssd} dataset.}
\label{fig:add2}
\end{figure*}

\begin{figure*}[hbt!]
\begin{center}
\scalebox{0.95}{
\includegraphics[width=1\textwidth]{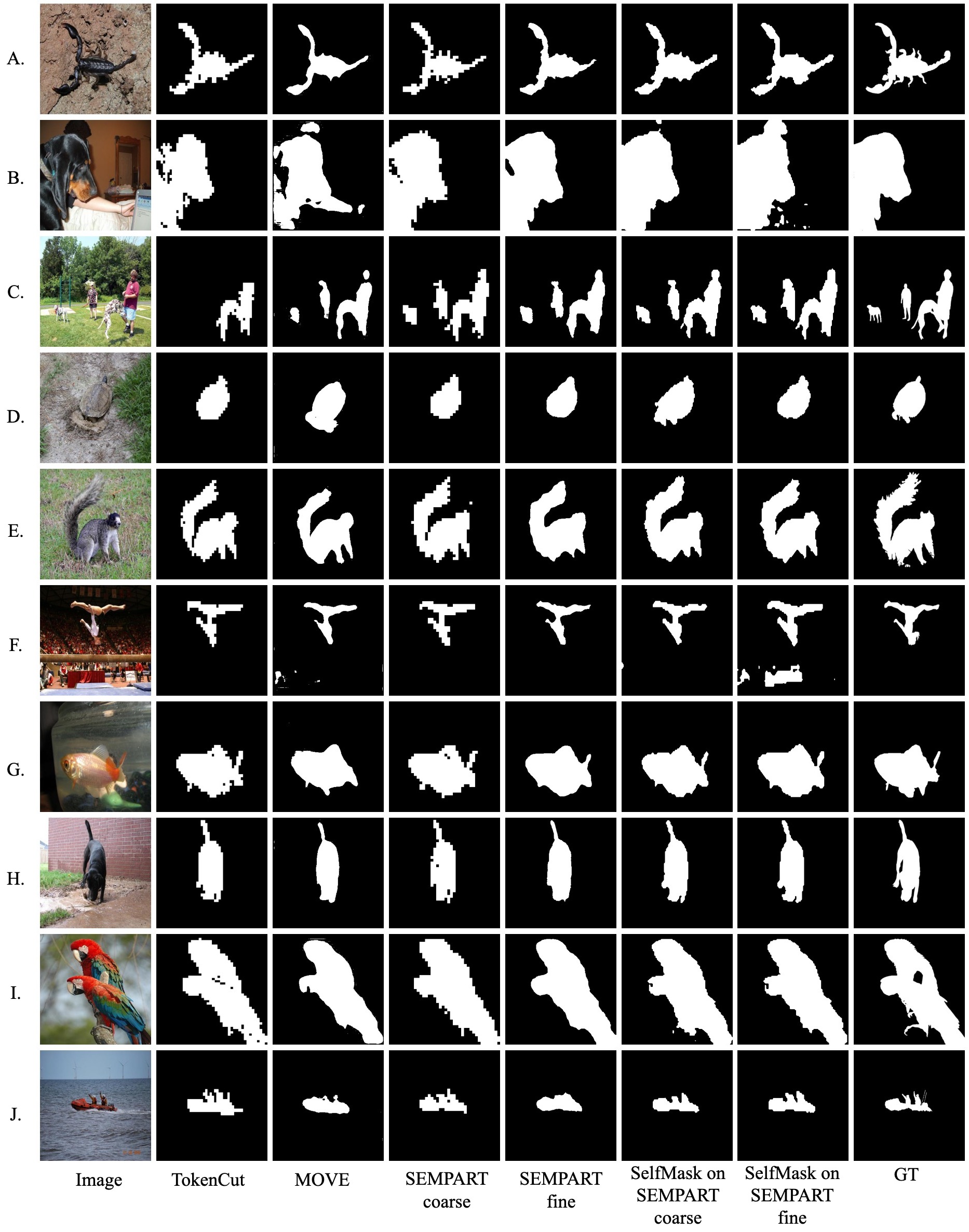}}
\end{center}
\caption{Additional examples on the DUTS-TE~\cite{duts} dataset.}
\label{fig:add3}
\end{figure*}

\noindent {\bf Additional results.}
\Figureref{add1}, \Figureref{add2} and \Figureref{add3} present additional results for both \textsc{Sempart}-coarse and -fine as well as also training \textsc{SelfMask}+\textsc{Sempart}-coarse and -fine as compared to TokenCut, MOVE, and the ground truth. The performance metrics in \Tableref{main_results} indicate that the average performance of additionally training \textsc{SelfMask} on \textsc{Sempart} as pseudo masks results in an improvement of 3\% and 3.5\% in IoU and $\max \text{F}_\beta$ respectively for the DUT-OMRON dataset. At the same time, the gains are debatable for DUTS-TE and, in particular, for ECSSD, for which the performance deteriorates for the \textsc{SelfMask} variant. 

Across \Figureref{add1}, \Figureref{add2}, and \Figureref{add3}, the superiority of \textsc{Sempart} over MOVE and TokenCut is a prevalent trend. As also seen previously in \Figureref{short}, TokenCut, which is optimized on a per image basis, not only results in {\it coarse masks} that do not capture several high-frequency details but can also select the incorrect object more often than its counterparts (see \Figureref{add1} (I)) as well as under select the salient region (see \Figureref{add1} (D, H), \Figureref{add3} (C)).

On the other hand, MOVE outperforms TokenCut by generating more accurate and high-resolution masks based on the perception of {\it movability} of foreground objects. This heuristic outperforms previous state-of-the-art significantly, as demonstrated in \cite{move}. However, we find that in addition to being noisy around the edges in most examples, it exhibits noisy artifacts both inside (see \Figureref{add1} (G), \Figureref{add2} (A, F), \Figureref{add3} (B)) and outside (see \Figureref{add1} (I), \Figureref{add2} (E), \Figureref{add3} (B, F))the visually salient regions. For the most part, MOVE can identify at least one of the salient objects. However, it seems likely that this heuristic also results in the over-selection of artifacts distinctly separated from the key salient object(s).

Compared to TokenCut and recent state-of-the-art MOVE, our method \textsc{Sempart} and its \textsc{SelfMask} variants signify a superior heuristic for unsupervised image bi-partitioning and a significantly better overlap with the ground truth saliency masks across all datasets. We also observe that the {\it fine mask} captures high-frequency details more accurately, especially at image boundaries than the corresponding jointly inferred {\it coarse mask}. The joint optimization involved in the \textsc{Sempart} architecture is valuable towards image bi-partitioning without involving any post-inference processing. Therefore the inference times are a fraction of its counterparts and comparable with other methods that also learn a segmentation model, such as MOVE. 

\begin{figure*}[hbt!]
\begin{center}
\scalebox{0.9}{
\includegraphics[width=1\textwidth]{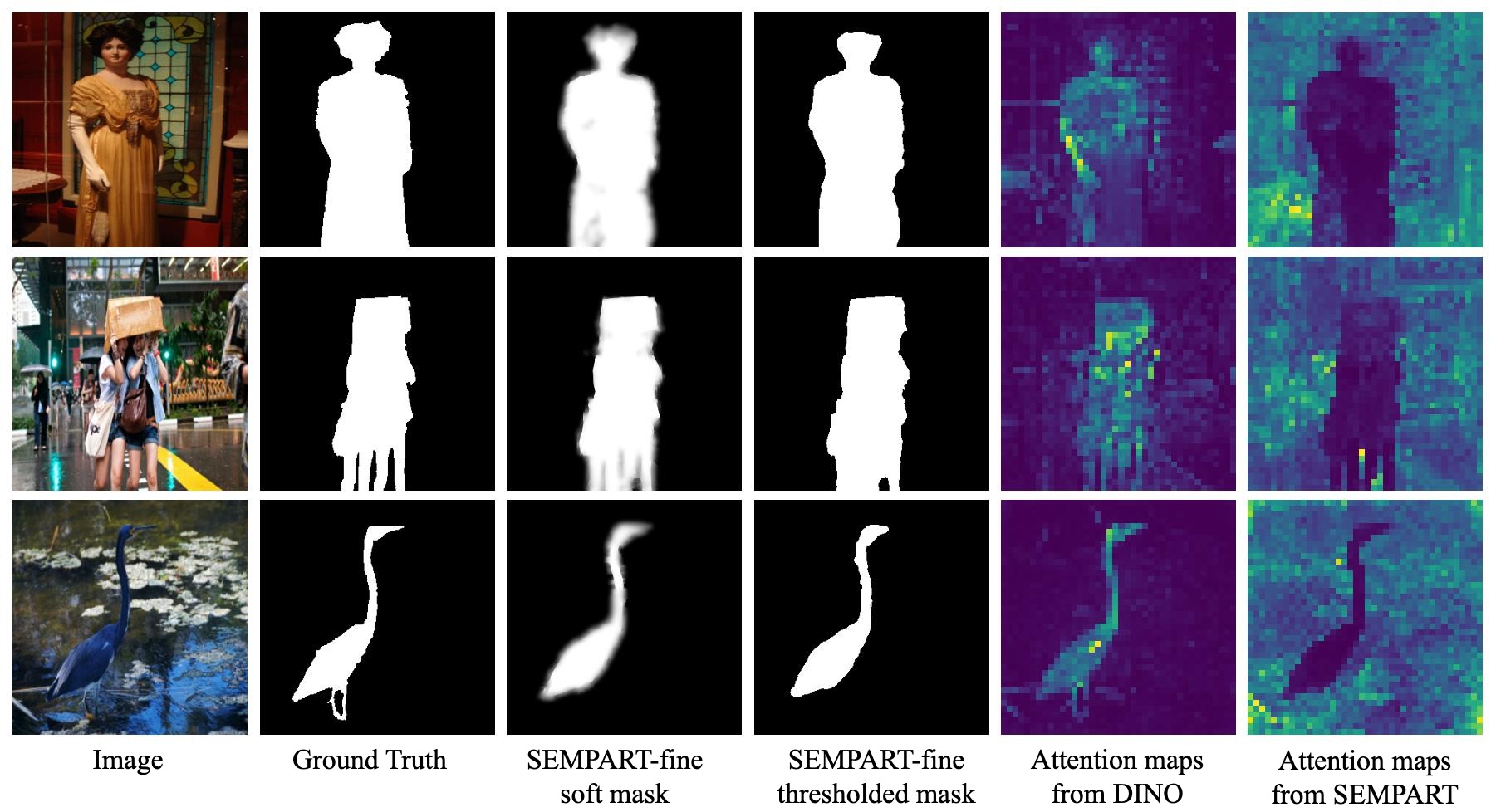}}
\end{center}

\caption{Attention map of the transformer encoder \texttt{[CLS]} token. The \textsc{Sempart} attention map aligns with the background.}

\label{fig:attnmap}
\end{figure*}

\vspace{1mm}
\noindent {\bf Attention map.}
The \textsc{TransformerEncoder} in \eqref{trfenc} is further elaborated in \Figureref{sempartarch} (a). To get a better understanding of the reasoning process of \textsc{Sempart}, we have looked at the average attention map across both heads for the \texttt{[CLS]} token of the \textsc{TransformerEncoder} in \Figureref{attnmap}. Interestingly we find that although the output of the \textsc{TransformerEncoder} for this particular token is discarded (see \Figureref{sempartarch} (a)), the corresponding attention map is insightful. This is because the \texttt{[CLS]} token is attended to by the remaining $40\times 40$ patch tokens for generating $\widetilde{F}$ in \eqref{trfenc}. Therefore, the underlying \texttt{[CLS]} embeddings get leveraged for the $\widetilde{F}$ output. In particular, the attention map resonates with the background\footnote{\cite{found} adopted a heuristic that expands the mask from background seeds located first.}. It reflects the clear distinction between an image's salient and non-salient regions. On the other hand, the DINO \texttt{[CLS]} token attention maps appear to attend to the foreground regions.

\section{Ethical aspects}
\label{app:ethics}
We benchmark our approach using publicly available datasets \cite{duts,dutomron,ecssd,voc07,voc12,cocodataset}. Although our approach infers unsupervised partitions of images, \textsc{Sempart} still inherits biases present in DINO~\cite{dino}, which was trained on ImageNet~\cite{imagenet} without labels and in a self-supervised manner.

\section{Future applications}
\label{app:future}
The merits of \textsc{Sempart} in generating high-quality masks at multiple resolutions can be particularly effective when applied to 
class-aware object detection, such as in \cite{namedmask}. More generally, \textsc{Sempart} can also help improve search and recommendation systems~\cite{search} in applications where users seek to retrieve images of specific objects with the underlying assumption that the object under consideration will likely be prominent and in the foreground. 


\end{document}